\documentclass[10pt,twocolumn,letterpaper]{article}

\usepackage[pagenumbers]{cvpr} 

\usepackage{graphicx}
\usepackage{amsmath}
\usepackage{amssymb}
\usepackage{booktabs}
\usepackage{mathalias}
\usepackage{multirow}
\usepackage{bbm}
\usepackage{caption}
\usepackage[table]{xcolor}
\usepackage{color}
\usepackage{pifont}
\usepackage{pifont}
\usepackage{algorithm}
\usepackage{listings}
\usepackage{placeins}
\usepackage[accsupp]{axessibility} 

\newcommand{\name}{CaSSLe}

\newcommand{\CC}[1]{\cellcolor{#1}}
\definecolor{ftcolor}{rgb}{1,1,1} 
\definecolor{baselinecolor}{rgb}{1,1,1}
\definecolor{contrcolor}{rgb}{1.0, 0.98, 0.8}
\definecolor{predcolor}{rgb}{0.74, 0.83, 0.9}
\definecolor{decorrcolor}{rgb}{0.98, 0.85, 0.87} 
\definecolor{knowcolor}{rgb}{0.82, 0.94, 0.75}
\definecolor{offlinecolor}{rgb}{1,1,1}
\definecolor{firsttaskcolor}{rgb}{0.97, 0.97, 0.97}
\definecolor{azure(colorwheel)}{rgb}{0.0, 0.5, 1.0}

\newcommand*\samethanks[1][\value{footnote}]{\footnotemark[#1]}

\usepackage[pagebackref,breaklinks,colorlinks=true,allcolors=blue]{hyperref}

\usepackage[capitalize]{cleveref}
\crefname{section}{Sec.}{Secs.}
\Crefname{section}{Section}{Sections}
\Crefname{table}{Table}{Tables}
\crefname{table}{Tab.}{Tabs.}

\def\cvprPaperID{****} 
\def\confName{****}
\def\confYear{****}

\newcommand{\mybullet}{\raisebox{1.5pt}{\scriptsize $\blacktriangleright$}}

\begin{document}

\title{Self-Supervised Models are Continual Learners}

\author{Enrico Fini\thanks{\scriptsize{Enrico Fini and Victor G. Turrisi da Costa contributed equally.}} \,$^{\textcolor{azure(colorwheel)}{1,2}}$ \quad Victor G. Turrisi da Costa\samethanks \, $^{\textcolor{azure(colorwheel)}{1}}$ \quad Xavier Alameda-Pineda$^{\textcolor{azure(colorwheel)}{2}}$\\
Elisa Ricci$^{\textcolor{azure(colorwheel)}{1,3}}$ \quad Karteek Alahari$^{\textcolor{azure(colorwheel)}{2}}$ \quad Julien Mairal$^{\textcolor{azure(colorwheel)}{2}}$\vspace{5px}\\
\normalsize{$^{\textcolor{azure(colorwheel)}{1}}$ University of Trento \quad $^{\textcolor{azure(colorwheel)}{2}}$ Inria\thanks{\scriptsize{Univ. Grenoble Alpes, CNRS, Grenoble INP, LJK, 38000 Grenoble, France.}} \quad $^{\textcolor{azure(colorwheel)}{3}}$ Fondazione Bruno Kessler}
}
\maketitle

\begin{abstract}
Self-supervised models have been shown to produce comparable or better visual representations than their supervised counterparts when trained offline on unlabeled data at scale. However, their efficacy is catastrophically reduced in a Continual Learning (CL) scenario where data is presented to the model sequentially. In this paper, we show that self-supervised loss functions can be seamlessly converted into distillation mechanisms for CL by adding a predictor network that maps the current state of the representations to their past state. This enables us to devise a framework for Continual self-supervised visual representation Learning that (i) significantly improves the quality of the learned representations, (ii) is compatible with several state-of-the-art self-supervised objectives, and (iii) needs little to no hyperparameter tuning. We demonstrate the effectiveness of our approach empirically by training six popular self-supervised models in various CL settings. Code: \href{https://github.com/DonkeyShot21/cassle}{\texttt{github.com/DonkeyShot21/cassle}}.
\end{abstract}\

\vspace{-5pt}
\section{Introduction}
\label{sec:intro}
\vspace{-2pt}
During the last few years, self-supervised learning (SSL) has become the most popular paradigm for unsupervised visual representation learning \cite{caron2020unsupervised, caron2021emerging, chen2020simple, he2020momentum, grill2020bootstrap, zbontar2021barlow, bardes2021vicreg, chen2020improved}. Indeed, under certain assumptions (\textit{e.g.}, offline training with large amounts of data and resources), SSL methods are able to extract representations that match the quality of representations obtained with supervised learning, without requiring annotations. However, these assumptions do not always hold in real-world scenarios, \textit{e.g.}, when new unlabeled data are made available progressively over time.
In fact, in order to integrate new knowledge into the model, training needs to be repeated on the whole dataset, which is impractical, expensive, and sometimes even impossible when old data is not available. This issue is exacerbated by the fact that SSL models are notoriously computationally expensive to train.

\begin{figure}[t]
\centering
\vspace{-5pt}
\includegraphics[width=0.99\columnwidth]{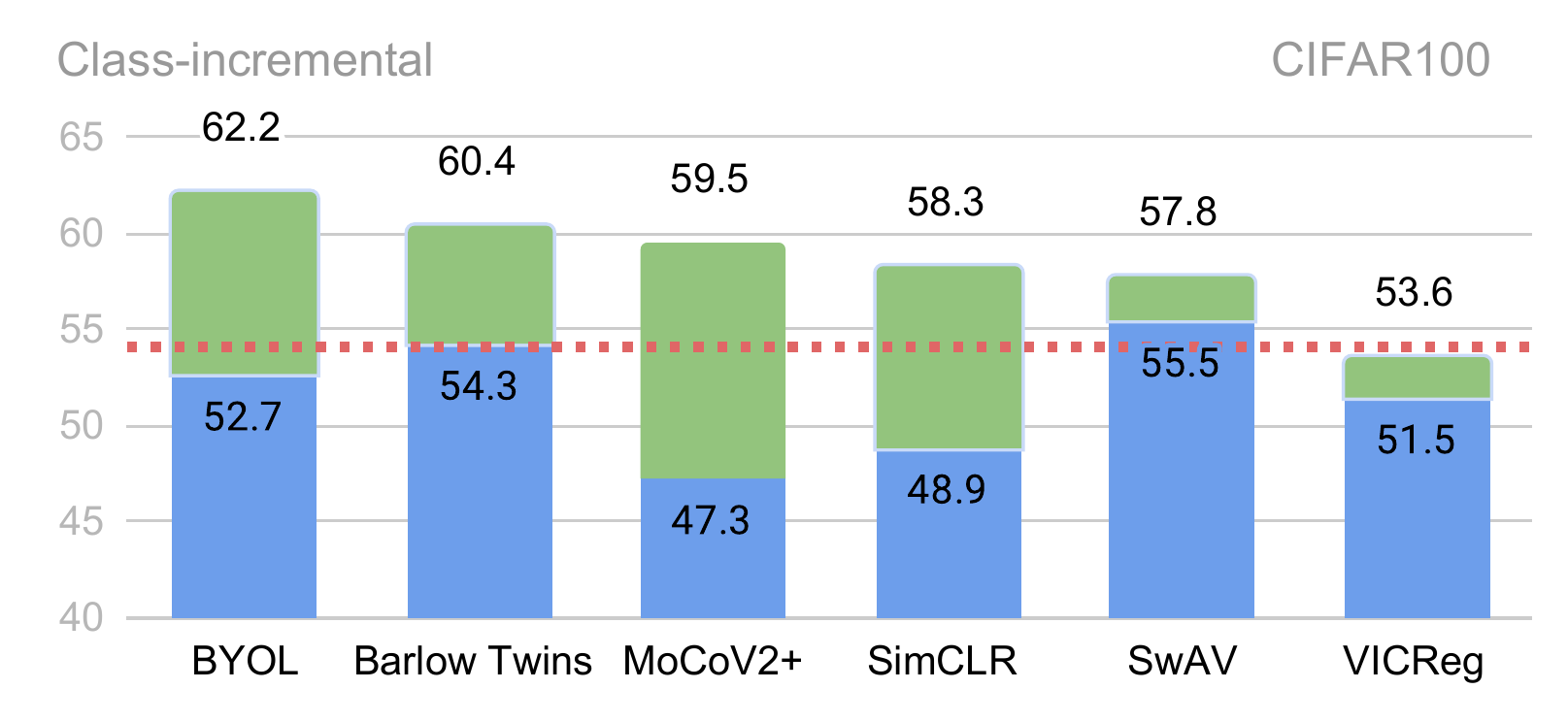}
\vspace{-12pt}
\includegraphics[width=0.99\columnwidth]{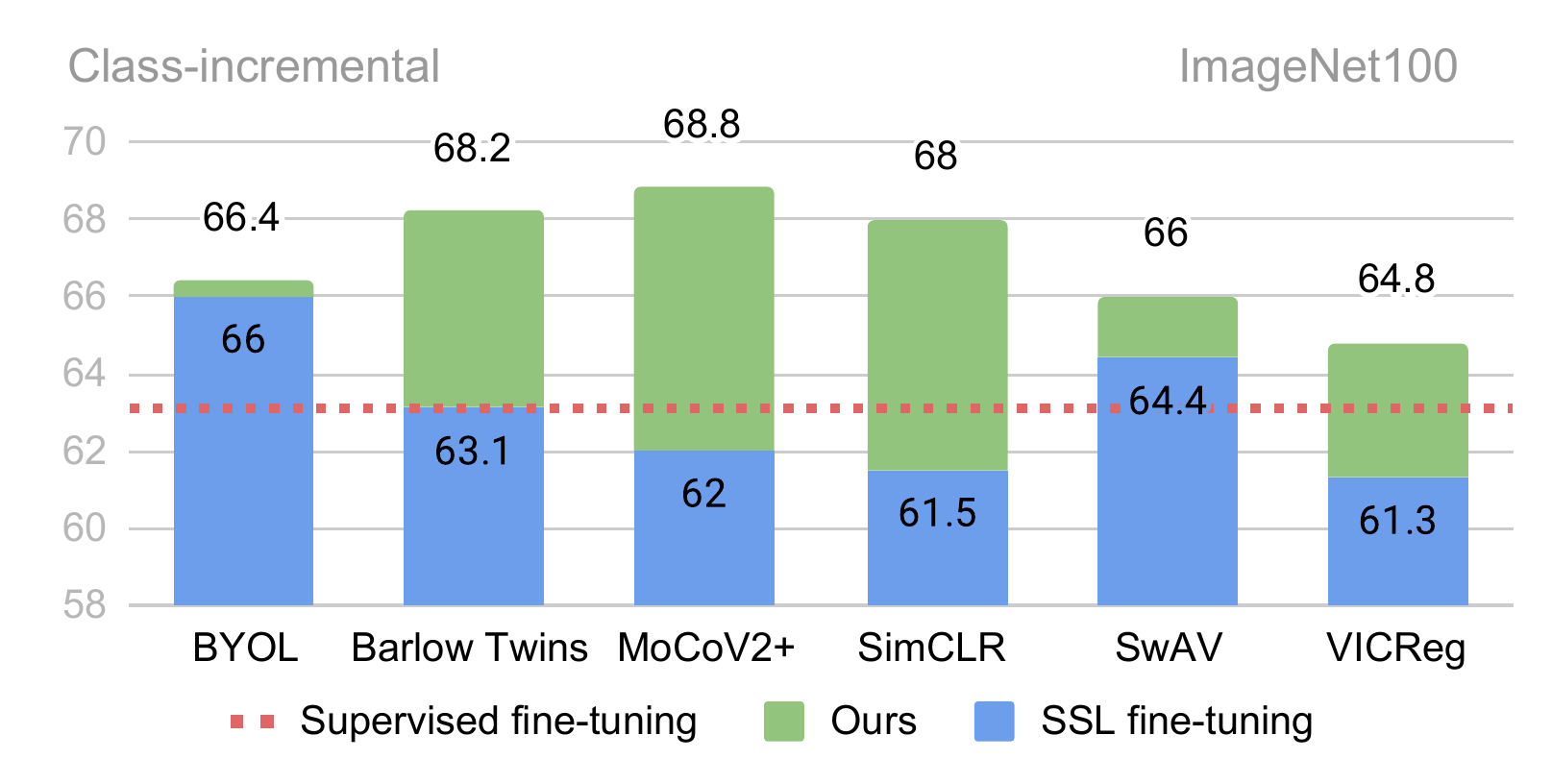}
\caption{Linear evaluation accuracy of representations learned with different self-supervised methods on class-incremental CIFAR100 and ImageNet100. In blue the accuracy of SSL fine-tuning, in green the improvement brought by \name{}. The red  dashed line is the accuracy attained by supervised fine-tuning.}
\vspace{-15pt}
\label{fig:teaser}
\end{figure}

Continual learning (CL) studies the ability of neural networks to learn tasks sequentially. Prior art in the field focuses on mitigating {catastrophic forgetting}~\cite{Mccloskey89,French99, Goodfellow13,de2019continual}. Common benchmarks in the CL literature evaluate the discriminative performance of classifiers learned \emph{with supervision} from non-stationary distributions. 
In this paper, we tackle the same forgetting phenomenon in the context of SSL. 
Unsupervised representation learning is indeed appealing for sequential learning since it does not require human annotations, which are particularly hard to obtain when new data is generated on-the-fly. This setup, called Continual Self-Supervised Learning (CSSL), is surprisingly under-investigated in the literature.

In this work, we propose \name{}, a simple and effective framework for CSSL of visual representations based on the intuition that SSL models are intrinsically capable of learning continually, and that SSL losses can be seamlessly converted into distillation losses. Our key idea is to train the current model to predict past representations with
a prediction head, thus encouraging it to remember past knowledge. \name{} has several favourable features: (i) it is compatible with popular state-of-the-art SSL loss functions and architectures, (ii) it is simple to implement, and (iii) it does not require any additional hyperparameter tuning with respect to the original SSL method. Our experiments demonstrate that SSL methods trained continually with \name{} significantly outperform all the related methods (CSSL baselines and several methods adapted from supervised CL).

We also perform a comprehensive analysis of the behavior of six popular SSL methods in diverse CL settings (\ie, class, data, and domain incremental). We provide empirical results on small (CIFAR100), medium (ImageNet100), and large (DomainNet) scale datasets. 
Our study sheds new light on interesting properties of SSL methods that emerge when learning continually.
Among other findings, we discover that, in the class-incremental setting, SSL methods typically approach or outperform supervised learning (see Fig.\ref{fig:teaser}), while this is not generally true for other settings (data-incremental and domain-incremental) where supervised learning still shows a sizeable advantage.

\vspace{-5pt}
\section{Related Work}
\label{sec:related}
\vspace{-8pt}
\noindent\textbf{Self-Supervised Learning.} 
Recent SSL approaches have shown performance comparable to their supervised learning equivalents~\cite{caron2020unsupervised, caron2021emerging, chen2020simple, he2020momentum, grill2020bootstrap, zbontar2021barlow, bardes2021vicreg, chen2020improved}. In a nutshell, most of these methods use image augmentation techniques to generate correlated views (positives) from a sample, and then learn a model that is invariant to these augmentations by enforcing the network to output similar representations for the positives. Initially, contrastive learning, based on instance discrimination~\cite{wu2018unsupervised} using noise-contrastive estimation~\cite{gutmann2010noise, oord2018representation}, was a popular strategy~\cite{chen2020simple, he2020momentum}. However, this learning paradigm 
requires large batch sizes or memory banks. A few methods that use a negative-free cosine similarity loss~\cite{grill2020bootstrap, chen2021exploring} have addressed such issues.

Concurrently, clustering-based methods (SwAV~\cite{caron2020unsupervised}, DeepCluster v2~\cite{caron2020unsupervised, caron2018deep} and DINO~\cite{caron2021emerging}) have also been proposed. They do not operate on the features directly, and instead compare positives through a cross-entropy loss using cluster prototypes as a proxy. Redundancy reduction-based methods have also been popular\cite{ermolov2021whitening, zbontar2021barlow, bardes2021vicreg}. Among them, BarlowTwins \cite{zbontar2021barlow} considers an objective function measuring the cross-correlation matrix between the features, and VicReg\cite{bardes2021vicreg} uses a mix of variance, invariance and covariance
regularizations. Methods such as~\cite{dwibedi2021little} have explored the use of nearest-neighbour retrieval and divide and conquer~\cite{tian2021divide}. However, none of these works studied the ability of SSL methods to learn continually and adaptively.

\noindent\textbf{Continual Learning.} A plethora of methods have been developed to counteract catastrophic forgetting~\cite{kirkpatrick2017overcoming, Rusu16progressive, shin2017continual, Lopez-Paz17, Chaudhry19, serra2018overcoming, chaudhry2018riemannian, Aljundi17, Zenke17, buzzega2020dark, fini2020online, douillard2020podnet, wu2019large, castro2018end, rebuffi2017icarl, hou2019learning, prabhu2020gdumb, Li17learning, ostapenko2019learning, cha2021co2l, Robins95}. Following~\cite{de2019continual}, these works can be organized into three macro-categories: replay-based~\cite{ostapenko2019learning, Robins95, rebuffi2017icarl, buzzega2020dark, Chaudhry19, Lopez-Paz17}, regularization-based~\cite{fini2020online, Li17learning, shin2017continual, kirkpatrick2017overcoming, Zenke17, Aljundi17, castro2018end, douillard2020podnet, hou2019learning, chaudhry2018riemannian, wu2019large, cha2021co2l}, and parameter isolation
methods~\cite{Rusu16progressive, serra2018overcoming}. All these works evaluate the effectiveness of CL methods using a linear classifier learned
sequentially over time. However, this evaluation does not reflect an important aspect, \textit{i.e.}, the internal dynamics of the hidden representations. Moreover, most CL methods tend to rely on supervision in order to mitigate catastrophic forgetting. A few of them can be adapted for the unsupervised setting, although their effectiveness is greatly reduced (see discussion in Sec.~\ref{sec:cassowary}, Sec.~\ref{sec:experiments} and the supplementary material). 

Works such as~\cite{rao2019continual, achille2018life, smith2019unsupervised} laid the foundations of unsupervised CL, but their studies are severely limited to digit-like datasets, \emph{e.g.}, MNIST and Omniglot, and the proposed methods are unfit for large-scale scenarios. Recently, \cite{gallardo2021self, caccia2021special} explored self-supervised pretraining for supervised continual learning with online and few-shot tasks, and \cite{cha2021co2l} presented a supervised contrastive CL approach. Two concurrent works~\cite{lin2021continual, madaan2021rethinking} have also attempted to address CSSL recently. The former~\cite{lin2021continual} extends~\cite{cha2021co2l} to the unsupervised setting, but is specifically designed for contrastive SSL, such as~\cite{chen2020simple,he2020momentum}, and lacks generalizability to other popular SSL paradigms. The latter~\cite{madaan2021rethinking} is also limited as it only shows small-scale experiments in the class-incremental setting and considers just two SSL methods. In contrast, we present a general framework for CSSL with superior performance, conduct large-scale experiments on three challenging settings, thereby presenting a deeper analysis of CSSL.
\vspace{-10pt}
\section{Preliminaries}
\label{sec:preliminaries}
\vspace{-7pt}

\noindent\textbf{Self-Supervised Learning.}
The training procedure of several state-of-the-art SSL methods \cite{zbontar2021barlow,chen2020simple,he2020momentum,caron2020unsupervised,caron2021emerging,bardes2021vicreg, dwibedi2021little,grill2020bootstrap} can be summarized as follows. Given an image $\xvect$ in a batch sampled from a distribution $\mathcal{D}$, two correlated views $\xvect^A$ and~$\xvect^B$ are extracted by applying stochastic image augmentations, such as random cropping, color jittering and horizontal flipping. View $\xvect^A$ is fed to an encoder $f_\theta = f_p \circ f_b$, which is parametrized by $\theta$ and has a backbone $f_b$ and a projection head $f_p$, that  extracts feature representations $\zvect^A = f_\theta(\xvect^A)$. Similarly, $\xvect^B$ is forwarded into the same networks, or possibly copies thereof, updated with exponential moving average (EMA), to obtain the representation~$\zvect^B$. A loss function $\mathcal{L}_{SSL}$ is applied to these representations to learn the parameters $\theta$ as follows:
\begin{equation}
\label{eq:ssl_obj}
    \underset{\theta}{\operatorname{argmin}} \,\, \mathbb{E}_{\xvect \sim \mathcal{D}}\left[\mathcal{L}_{SSL}\left(\zvect^A, \zvect^B\right)\right].
\end{equation}
More details on the implementation of $\mathcal{L}_{SSL}$ are provided in Sec.~\ref{sec:compatibility} and Tab.~\ref{tab:methods}. This procedure turns out to be extremely powerful at extracting visual representations from large unlabeled datasets. The intuition behind the success of these models is that they learn to be invariant to augmentations. Importantly, augmentations are hand-crafted in a way that the two views $\xvect^A$ and $\xvect^B$ contain roughly the same semantics as $\xvect$, but their overall appearance (geometry, colors, resolution, \etc) is different. This forces the model map images with the same semantics to similar regions of the feature space. Interestingly, these augmentations are much stronger, \ie, they distort the image more, than augmentations commonly used to train supervised models.

\noindent\textbf{Continual Learning.}
The CL problem focuses on training models such as deep neural networks from non-stationary data distributions. More formally, this involves a network $f^{\prime}_{\theta^{\prime}} = f^{\prime}_c \circ f^{\prime}_b $ with parameters $\theta^{\prime}$, backbone $f^{\prime}_b$ and a classifier $f^{\prime}_c$, that learns from an ordered set of tasks $\{1,\dots, T\}$, each exhibiting a different data distribution $\mathcal{D}_t$. Usually, an image $\xvect$ sampled i.i.d.\ from $\mathcal{D}_t$ is processed by $f^{\prime}$ that predicts a probability distribution $\pvect$ over the set of classes $\mathcal{Y}_t$. The objective is to find parameters $\theta^{\prime}$ such as:
\begin{equation}
\label{eq:cl_obj}
\underset{\theta^{\prime}}{\operatorname{argmin}} \sum_{t=1}^{T} \mathbb{E}_{(\xvect, \yvect) \sim \mathcal{D}_{t}}\left[\mathcal{L}_{CL}\left(\pvect, \yvect\right)\right],
\end{equation}
where, in most cases, $\mathcal{L}_{CL}$ is the cross-entropy loss. However, during task $t$, the previous data distribution $\mathcal{D}_{t-1}$ is not available and therefore Eq.~(\ref{eq:cl_obj}) cannot be minimized directly. Current research focuses on approximating $\theta^{\prime}$ using indirect approaches. Some of them~\cite{Li17learning, douillard2020podnet} are based on knowledge distillation~\cite{hinton2015distilling}, \ie, transferring knowledge from one network to another by forcing them to produce the same outputs. We will discuss the applicability of distillation methods in CSSL in Sec.~\ref{sec:cassowary}.

\vspace{-5pt}
\section{Continual Self-Supervised Learning}
\label{sec:cssl}
In this paper, we tackle the problem of Continual Self-Supervised Learning 
as an extension of both SSL and CL.
In practice, a CSSL experiment starts with the first task, where the model is trained as per the specific self-supervised method that it implements, with no difference from offline training. Subsequent tasks are then presented to the model sequentially, and the data from the previous tasks are discarded.
No labels are provided during this training phase. For the sake of simplicity and since we are exploring a new, challenging setting, we assume task boundaries to be provided to the model.
More formally, the CSSL objective is to learn a strong feature extractor that is invariant to augmentations on all tasks. Following the notation introduced in Sec.~\ref{sec:preliminaries},  we define: 
\begin{equation}
\label{eq:continual_erm}
\underset{\theta}{\operatorname{argmin}} \sum_{t=1}^{T} \mathbb{E}_{\xvect \sim \mathcal{D}_{t}}\left[\mathcal{L}_{SSL}\left(\zvect^A, \zvect^B\right)\right].
\end{equation}
Note the absence of labels $\yvect$ when sampling from $\mathcal{D}_t$, the summation over the set of tasks inherited from Eq.~(\ref{eq:cl_obj}) and the SSL loss function in Eq.~(\ref{eq:ssl_obj}). The expectation is approximated using stochastic gradient descent on minibatches.

\noindent\textbf{Evaluation.} After each task, it is possible (for evaluation purposes) to train a linear classifier on top of the obtained backbone $f_b$. With this linear classifier we report accuracy on the test set. This protocol is compatible with standard CL metrics, as shown in Sec.~\ref{sec:protocol}.
We explore three CSSL settings in our work.\\
\mybullet\ \textbf{Class-incremental}: each task $t$ is represented by a dataset $D_t \sim \mathcal{D}_t$ containing images that belong to a set of classes $\mathcal{Y}_t$ such that $\mathcal{Y}_t \cap \mathcal{Y}_s = \varnothing$ for each other task $s \neq t$. Note that the class labels are only used for splitting the dataset and they are unknown to the model. In practice, the set of classes in the dataset are shuffled and then partitioned into $T$ tasks. Each task contains the same number of classes.\\
\mybullet\ \textbf{Data-incremental}: each task $t$ contains a set of images $D_t$ such that $D_t \cap D_s = \varnothing$ for each other task $s \neq t$. No additional constraints are imposed on the classes. In practice, the whole dataset is shuffled and then partitioned into $T$ tasks. Each task can potentially contain all the classes. \\
\mybullet\ \textbf{Domain-incremental}: each task $t$ contains a set of images $D_t$ drawn from a different domain. We assume that the set of classes $\mathcal{Y}_t$ in each dataset remains the same for all tasks but the data distribution changes, as if the data were collected from different sources. 
\vspace{-5pt}
\section{The \name{} Framework}
\label{sec:cassowary}
\vspace{-5pt}
We now introduce ``\name{}'', our framework for 
continual self-supervised learning of visual representations and detail its compatibility with several SSL methods.
\vspace{2pt}

\noindent{\bf Distillation in CSSL.} From a supervised CL perspective, the concept of invariance is interesting. Here, we would like to learn representations of previously-learned semantic concepts that are invariant to the state of the model's parameters. 
Indeed, this idea was investigated in prior works~\cite{douillard2020podnet, hou2019learning} that leverage knowledge distillation for CL. However, such approaches are only  mildly effective in a CSSL scenario, as we show in Sec.~\ref{sec:experiments}. 
We believe this is due to CSSL being fundamentally different from supervised CL. In CSSL, we aim to extract the best possible representations that can be subsequently reused in a variety of tasks, and maximize the linear separability of features at the end of the CL phase. Thus, the linear classifier does not benefit much from the stability of the representations. Also, forcing the representations not to change may prevent the model from learning new concepts. 
This is especially critical for SSL methods for two reasons: (i) the performance of the models improve substantially with longer training, implying that the representations continue to get refined, and (ii) they exhibit different losses and feature normalizations that might interfere with distillation and vice-versa (\eg, BarlowTwins uses standardization while~\cite{douillard2020podnet, hou2019learning} use $l2$-normalization). Nonetheless, the features still need to be informative of previous tasks to maximize the separability of the old distribution but the current state might be too different from the previous one making comparing representations complicated.

\vspace{2pt}
\noindent{\bf Distillation through SSL losses.}
Our framework, shown in Fig.~\ref{fig:\name{}}, is based on the following ideas: (i)~a predictor network that maps the current state of the representations to their past state, by leveraging a distillation through time strategy that satisfies both stability and plasticity principles, and (ii)~a family of adaptable distillation losses inherited from the SSL literature that solves the issue of having different objectives interfering with each other.

\begin{figure}[t]
\centering
\includegraphics[width=0.85\columnwidth]{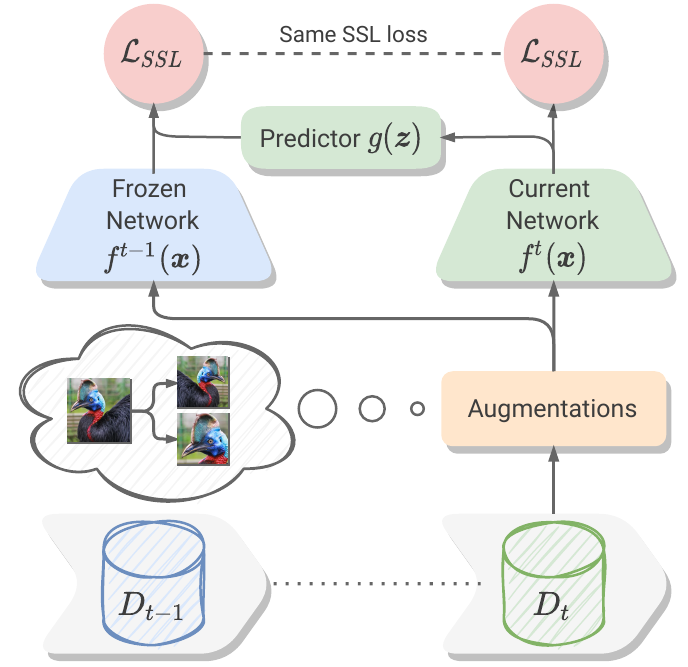}
\vspace{-8pt}
\caption{Overview of the \name{} framework.}
\label{fig:\name{}}
\vspace{-11pt}
\end{figure}

When a new task is received, we start by making a copy of the current model. This copy does not require gradient computation and will not be updated. We call this the \textit{frozen encoder} $f^{t-1}$. As soon as an image $\xvect \in D_t$ is available we apply our stochastic image augmentations and extract its features $\zvect = f^t(\xvect)$. In addition, we also use the frozen encoder to extract another feature vector $\bar{\zvect} = f^{t-1}(\xvect)$. Now, our goal is to ensure that $\zvect$ contains at least as much information as (and ideally more than) $\bar{\zvect}$. Instead of enforcing the two feature vectors to be similar, and hence discouraging the new model from learning new concepts, we propose 
to use a predictor network $g$ to project the representations from the new feature space to the old one. If the predictor is able to perfectly map from one space to the other, then it implies that $\zvect$ is at least as powerful as $\bar{\zvect}$.

We are now ready to perform distillation, but which is the most appropriate distillation loss? 
Since we want the representations produced by $g$ to be invariant to the state of the model, we propose to use the same SSL loss used to simulate invariance to augmentations. Empirically, we verify that this choice reduces interference and minimizes the need for hyperparameter tuning. We can hence write a generic distillation loss by reusing the definition of $\mathcal{L}_{SSL}$:
\begin{equation}
    \mathcal{L}_D(\zvect, \bar{\zvect}) = \mathcal{L}_{SSL} (g(\zvect), \bar{\zvect}).
\end{equation}
Note that $\bar{\zvect}$ is always detached from the computational graph, such that the frozen encoder does not receive any gradient, and the gradient only flows through the predictor $g$, as prescribed in~\cite{chen2021exploring}. On the one hand, if training  converges and $\mathcal{L}_D$ is minimized, the features predicted by $g$ will likely be quasi-invariant to the state of the model, which satisfies the stability principle. On the other hand, the current encoder is less bound to its previous state, hence representations $\zvect$ can be more plastic. The loss can be extended to multiple views by applying it to both representations, \textit{i.e.}, $\mathcal{L}_D(\zvect^A, \bar{\zvect}^A) + \mathcal{L}_D(\zvect^B, \bar{\zvect}^B)$, and also swapped distillation, \textit{e.g.}, $\mathcal{L}_D(\zvect^A, \bar{\zvect}^B)$ and vice-versa (see ablation in Tab.~\ref{tab:ablation}).

The final loss of an SSL method trained continually with the \name{} framework is given by:
\begin{equation}
\begin{aligned}
    \mathcal{L} &= \mathcal{L}_{SSL}(\zvect^A, \zvect^B) + \mathcal{L}_{D}(\zvect^A, \bar{\zvect}^A) \\
    &= \mathcal{L}_{SSL}(\zvect^A, \zvect^B)  +\mathcal{L}_{SSL} (g(\zvect^A), \bar{\zvect}^A).
\end{aligned}
\label{eq:\name{}}
\end{equation}
This loss can be made symmetric by applying it to both the views (swapping $A$ and $B$ in Eq.~(\ref{eq:\name{}})) and it can also be easily adapted for multi-crop~\cite{caron2020unsupervised}. Note that we do not use any hyperparameter to weight the importance of the distillation loss with respect to the SSL loss.

\vspace{-5pt}
\subsection{Compatibility of SSL methods with \name{}}
\label{sec:compatibility}
\vspace{-5pt}
The main difference among SSL methods is the loss function that they use. Following the notation defined in Sec.~\ref{sec:preliminaries}, and the loss functions in Tab.~\ref{tab:methods}, we now detail if and how SSL losses can be used in our \name{} framework. Full derivation of distillation losses is deferred to the supplementary material.

\noindent\textbf{InfoNCE-based} methods \cite{chen2020simple, he2020momentum} perform instance discrimination, where positive samples help to build invariance to augmentations. The negatives prevent the model from falling into degenerate solutions. The InfoNCE (\textit{a.k.a.} contrastive) loss can be written as in Eq.~(\ref{eq:infonce}), where subscript~$i$ is the index of a generic sample in the batch, $\operatorname{sim}$ is the cosine similarity and $\eta(i)$ is the set of negatives for sample $i$ in the current batch. 
Distilling knowledge with this loss is equivalent to performing instance discrimination of current task samples but in the feature space learnt in the past. Thus, the predictor $g$ learns to project samples from the present to the past space to maximize the distance with the negative samples, and the similarity with itself in the past.

\noindent\textbf{MSE-based} approaches~\cite{grill2020bootstrap, chen2021exploring} enforce consistency among positive samples and ignore the negatives. BYOL~\cite{grill2020bootstrap} uses a momentum encoder and SimSiam~\cite{chen2021exploring} performs a stop gradient operation to avoid degenerate solutions. Since the representations are $l2$-normalized, their loss (Eq.~\ref{eq:mse}) can be rewritten as the negative cosine similarity:  $-\operatorname{sim}(\qvect^A, \zvect^B) =-\frac{\qvect^A}{||\qvect^A||_2}\cdot\frac{\zvect^B}{||\zvect^B||_2}$,
where $\qvect^A = h(\zvect^A)$ and $h$ is a prediction head. The gradient is backpropagated only through the representations of the first argumentation. A special case of this family of methods is VICReg~\cite{bardes2021vicreg}, which uses a combination of multiple losses, where MSE acts as invariance term. Features are not $l2$-normalized in VICReg and its predictor is the identity function. In our framework, this loss encourages the model to predict the past state of representations without additional regularization.

\begin{table}[t]
\caption{Overview of state-of-the-art SSL methods and losses. In all tables, highlight colors are coded according to the type of loss.}
\label{tab:methods}
\vspace{-8pt}
\centering
\resizebox{0.99\columnwidth}{!}{
\captionsetup{type=table}
\begin{tabular}{m{1.9cm}ccc}
\toprule
\textbf{Methods} & \textbf{Loss} & \textbf{Equation} \\ 
\midrule
SimCLR~\cite{chen2020simple} & \CC{contrcolor} & \multirow{3}{*}{$-\log \frac{\exp \left(\operatorname{sim}\left(\zvect_i^A, \zvect_i^B\right) / \tau\right)}{\sum_{\zvect_j \in \eta(i)} \exp \left(\operatorname{sim}\left(\zvect_i^A, \zvect_{j}\right) / \tau\right)}$} & \hspace{-3mm} \multirow{3}{*}{$\refstepcounter{equation}(\theequation)\label{eq:infonce}$} \\
MoCo~\cite{he2020momentum} & \CC{contrcolor} \\
NNCLR~\cite{dwibedi2021little} & \CC{contrcolor} \multirow{-3}{*}{InfoNCE} \\\midrule

BYOL~\cite{grill2020bootstrap} &  \CC{predcolor} & \multirow{3}{*}{$-||\qvect^A -  \zvect^B||^2_2$} & \hspace{-3mm}\multirow{3}{*}{$\refstepcounter{equation}(\theequation)\label{eq:mse}$} \\
SimSiam~\cite{chen2021exploring} & \CC{predcolor}\\
VICReg~\cite{bardes2021vicreg} & \CC{predcolor}\multirow{-3}{*}{MSE}\\\midrule

SwAV~\cite{caron2020unsupervised} & \CC{knowcolor}  & \multirow{3}{*}{$-\sum_{d} \avect_{d}^B \log \frac{\exp \left(\operatorname{sim}\left( \zvect^A, \cvect_{d}\right) / \tau\right)}{\sum_k \exp \left(\operatorname{sim}\left(\zvect^A, \cvect_k\right) / \tau\right)} $} & \hspace{-3mm} \multirow{3}{*}{$\refstepcounter{equation}(\theequation)\label{eq:cross-entropy}$} \\
DCV2~\cite{caron2020unsupervised} & \CC{knowcolor}\\
DINO~\cite{caron2021emerging} & \CC{knowcolor}\multirow{-3}{*}{Cross-entropy} \\\midrule

Barlow Twins~\cite{zbontar2021barlow} &\CC{decorrcolor}  & \multirow{2}{*}{$\sum_{u}\left(1-\mathcal{C}_{u v}\right)^{2}+\lambda \sum_{u} \sum_{v \neq u} \mathcal{C}_{u v}^{2}$} & \hspace{-3mm} \multirow{2}{*}{$\refstepcounter{equation}(\theequation)\label{eq:cross-correlation}$}\\
VICReg~\cite{bardes2021vicreg} &\CC{decorrcolor}\multirow{-3}{*}{Cross-correlation} \\

\bottomrule
\end{tabular}
}
\vspace{-13pt}
\end{table}
\noindent\textbf{Cross-entropy-based.} Instead of simply enforcing invariance of the representations to augmentations, cluster prototypes $\mathbf{C} = \left\{\mathbf{c}_{1}, \ldots, \mathbf{c}_{K}\right\}$ are used as a proxy in these approaches, so that the model learns to predict invariant cluster assignments. Slight variations of this idea result in different methods: SwAV~\cite{caron2020unsupervised}, DeepClusterV2~\cite{caron2020unsupervised} and DINO~\cite{caron2021emerging}. Once a probability distribution over the prototypes is predicted, the cross-entropy loss (Eq.~\ref{eq:cross-entropy}) is used to compare the two views.
Features and cluster prototypes $\cvect$ are $l2$-normalized. The assignments $\avect^B$ can be calculated in several ways, \textit{e.g.}, k-means in DeepCluster, Sinkhorn-Knopp in SwAV and EMA in DINO. When employed as a distillation loss, cross-entropy encourages $g$ to predict the assignments generated by the frozen encoder with a set of frozen prototypes: $\avect^B = \frac{\exp \left(\operatorname{sim}\left( \bar{\zvect}^B, \cvect^{t-1}_{d}\right) / \tau\right)}{\sum_k \exp \left(\operatorname{sim}\left(\bar{\zvect}^B, \cvect^{t-1}_k\right) / \tau\right)}$, where $\mathbf{C}^{t-1} = \left\{\mathbf{c}^{t-1}_{1}, \ldots, \mathbf{c}^{t-1}_{K}\right\}$.

\noindent\textbf{Cross-correlation-based}. These methods use a different approach based on decorrelating the components of the feature space, \textit{e.g.}, Barlow Twins~\cite{zbontar2021barlow}, VICReg~\cite{bardes2021vicreg} and W-MSE~\cite{ermolov2021whitening}. For our analysis, we will mainly focus on Barlow Twins' implementation of this objective. Extensions to VICReg are left for future work. The cross-correlation based objective function is shown in Eq.~\ref{eq:cross-correlation}, 
where $\lambda$ is an hyperparameter to control the importance of the first and the second terms of the loss, and $\mathcal{C}_{u v} = \frac{\sum_{i} \zvect_{i, u}^{A} \zvect_{i, v}^{B}}{\sqrt{\sum_{i}\left(\zvect_{i, u}^{A}\right)^{2}}. \sqrt{\sum_{i}\left(\zvect_{i, v}^{B}\right)^{2}}}$ is the value of position $(u,v)$ of the cross-correlation matrix computed between the representations of the views along the batch dimension. Note that the representations here are mean centered along the batch dimension, such that each unit has mean output zero over the batch. Performing distillation with this loss has the additional effect of decorrelating the dimensions of the predicted features $g(\zvect^A)$.
\vspace{-5pt}
\section{Experiments}
\label{sec:experiments}
\vspace{-5pt}
\subsection{Experimental Protocol}
\label{sec:protocol}
\vspace{-5pt}
\paragraph{Evaluation Metrics.}
Following previous work~\cite{Lopez-Paz17}, we propose the following metrics to evaluate the quality of the representations extracted by our CSSL model:\\
\mybullet\ \textbf{Linear Evaluation Accuracy}: accuracy of a classifier trained on top of the backbone $f_b$ on all tasks (or a subset, \eg, 10\% of the data) or a downstream task. For class-incremental and data-incremental, we use the task-agnostic setting, meaning that at evaluation time we do not assume to know the task ID. For the domain-incremental setting, we perform both task-aware and task-agnostic evaluations (the latter is discussed in the supplementary material). To calculate the average accuracy we compute $A = \frac{1}{T} \sum_{i=1}^{T} A_{T, i},$ where $A_{j,k}$ is the linear evaluation accuracy of the model on task $k$ after observing the last sample from task $j$.\\
\mybullet\ \textbf{Forgetting}: a common metric in the CL literature, it quantifies how much information the model has forgotten about previous tasks. It is formally defined as: $F=\frac{1}{T-1} \sum_{i=1}^{T-1} \max _{t \in\{1, \ldots, T\}}\left(A_{t, i}-A_{T, i}\right)$.\vspace{1mm}\\
\mybullet\ \textbf{Forward Transfer}: measures how much the representations that we learned so far are helpful in learning new tasks, namely: $FT = \frac{1}{T-1} \sum_{i=2}^{T} A_{i-1, i}-R_{i}$ where $R_{i}$ is the linear evaluation accuracy of a random network on task $i$.

\noindent\textbf{Datasets.} We perform experiments on 3 datasets: {CIFAR100}~\cite{krizhevsky2009learning} (class-incremental), a 100-class dataset with 60k 32x32 colour images; {ImageNet100}~\cite{tian2020contrastive} (class- and data-incremental), 100-class subset of the ILSVRC2012 dataset with $\approx$130k images in high resolution (resized to 224x224); {DomainNet}~\cite{peng2019moment} (domain-incremental), a 345-class dataset containing roughly 600k high-resolution images (resized to 224x224) divided into 6 domains. We experiment with 5 tasks for the class- and data-incremental settings and with 6 tasks (one for each domain in DomainNet) in the case of domain-incremental. The supplementary material presents additional results with different number of tasks. For the domain-incremental setting, we order the domains in decreasing number of images.

\noindent\textbf{Implementation details.}
The SSL methods are adapted from \texttt{solo-learn}~\cite{turrisi2021sololearn}, an established SSL library, which is the main code base for all our experiments. 
The number of epochs per task is as follows: 500 for CIFAR100, 400 for ImageNet100, 200 for DomainNet. The backbone $f_b$ is a ResNet18~\cite{he2016deep}, with batch size 256. We use LARS~\cite{you2017large} for all our experiments. The offline version of each method, that serves as an upper bound, is trained for the same number of epochs as the continual counterpart for a fair comparison. All the results for offline upper bounds are obtained using the checkpoints provided in~\cite{turrisi2021sololearn}.  For some SSL methods, it was necessary to slightly increase the learning rate over the values provided by~\cite{turrisi2021sololearn} in order for the methods to fully convergence in the CSSL setting. Although tuning the hyperparameters might be beneficial in some settings, we do \textbf{not} perform any hyperparameter tuning for \name{}. We also neither change the parameters of the SSL methods, nor use a weight for the distillation loss (as per Eq.~(\ref{eq:\name{}})). 

\noindent\textbf{Baselines.}
Most of the CL methods require labels which makes them unsuitable for CSSL. However, a few works can be adapted for our setting with minimal changes. We choose baselines from three categories~\cite{de2019continual}: prior-focused regularization (EWC~\cite{kirkpatrick2017overcoming}), data-focused regularization (POD~\cite{douillard2020podnet}, Less-Forget~\cite{hou2019learning}), and rehearsal-based replay (ER~\cite{Robins95}, DER~\cite{buzzega2020dark}) methods. We also compare with two concurrent works that propose approaches for CSSL (LUMP~\cite{madaan2021rethinking}, Lin \etal~\cite{lin2021continual}). Finally, we do not consider methods based on VAEs~\cite{rao2019continual, achille2018life}, since they have been shown to yield poor performance on large scale. Details on how the baselines are selected, implemented and tuned for CSSL can be found in the supplementary material.

\begin{table}[t]
\caption{Comparison with state-of-the-art CL methods on CIFAR100 (5 tasks, class-incremental) using linear evaluation top-1 accuracy, forgetting and forward transfer.}
\label{tab:comp-with-sota}
\vspace{-7pt}
\centering
\scriptsize
\setlength{\tabcolsep}{2.2pt}
\captionsetup{type=table}
\begin{tabular}{lccccccccc}
\toprule
\multirow{2}[1]{*}{\textbf{Strategy}} & \multicolumn{3}{c}{SimCLR} & \multicolumn{3}{c}{Barlow Twins} & \multicolumn{3}{c}{BYOL}\\
\cmidrule(lr){2-4}\cmidrule(lr){5-7}\cmidrule(lr){8-10}
    & \textbf{A ($\uparrow$)} & \textbf{F ($\downarrow$)} & \textbf{T ($\uparrow$)} & \textbf{A ($\uparrow$)} & \textbf{F ($\downarrow$)} & \textbf{T ($\uparrow$)} & \textbf{A ($\uparrow$)} & \textbf{F ($\downarrow$)} & \textbf{T ($\uparrow$)} \\ 
\midrule
\CC{ftcolor}Fine-tuning & \CC{ftcolor}48.9 & \CC{ftcolor}1.0 & 33.5  & \CC{ftcolor}54.3 & \CC{ftcolor}0.4 & \CC{ftcolor}39.2  & \CC{ftcolor}52.7 & \CC{ftcolor}0.1 & 35.9 \\
\CC{baselinecolor}EWC~\cite{kirkpatrick2017overcoming} & \CC{baselinecolor}53.6 & \CC{baselinecolor}\textbf{0.0} & 33.3 & \CC{baselinecolor}56.7& \CC{baselinecolor}\textbf{0.2}  & \CC{baselinecolor}39.1 & 56.4 & \CC{baselinecolor}\textbf{0.0} & 39.9 \\  
\CC{baselinecolor}ER~\cite{Robins95} &\CC{baselinecolor}50.3 & \CC{baselinecolor}0.1 & 32.7 &\CC{baselinecolor}54.6 & \CC{baselinecolor}3.0 & \CC{baselinecolor}39.4 &\CC{baselinecolor}54.7 & \CC{baselinecolor}0.4 & 36.3\\  
\CC{baselinecolor}DER~\cite{buzzega2020dark} &\CC{baselinecolor}50.7 & \CC{baselinecolor}0.4 & 33.2 &\CC{baselinecolor}55.3 & \CC{baselinecolor}2.5 & \CC{baselinecolor}39.6 &\CC{baselinecolor}54.8 & \CC{baselinecolor} 1.1 & 36.7 \\  
\CC{baselinecolor}LUMP~\cite{madaan2021rethinking} &\CC{baselinecolor}52.3 & \CC{baselinecolor}0.3 & 34.5 &\CC{baselinecolor}57.8 & \CC{baselinecolor}0.3 & \CC{baselinecolor}41.0 & 56.4 & 0.2 & 37.9 \\ 
\CC{baselinecolor}Less-Forget\cite{hou2019learning} & \CC{baselinecolor}52.5 & \CC{baselinecolor}0.2 & 33.8 & \CC{baselinecolor}56.4 & \CC{baselinecolor}\textbf{0.2}   & \CC{baselinecolor}40.1 & \CC{baselinecolor}58.6 & \CC{baselinecolor}0.2 & 41.1 \\  
\CC{baselinecolor}POD\cite{douillard2020podnet} & \CC{baselinecolor}51.3 & \CC{baselinecolor}0.1 & 33.8  & \CC{baselinecolor}55.9 & \CC{baselinecolor}0.3   & \CC{baselinecolor}40.3 & \CC{baselinecolor}57.9 & \CC{baselinecolor}\textbf{0.0}  & 41.1  \\
\CC{baselinecolor}\name{}  & \CC{contrcolor}\textbf{58.3} & \CC{contrcolor}0.2 & \CC{contrcolor}\textbf{36.4} & \CC{decorrcolor}\textbf{60.4} & \CC{decorrcolor}0.4   & \CC{decorrcolor}\textbf{42.2} & \CC{predcolor}\textbf{62.2} & \CC{predcolor}\textbf{0.0} & \CC{predcolor}\textbf{43.6}   \\
                             \midrule
 \CC{offlinecolor}Offline & \CC{offlinecolor}65.8 & \CC{offlinecolor}-   & \CC{offlinecolor}- & \CC{offlinecolor}70.9   & \CC{offlinecolor}-  & \CC{offlinecolor}- & \CC{offlinecolor}70.5 & \CC{offlinecolor}-  & \CC{offlinecolor}-\\
\bottomrule
\end{tabular}
\captionsetup{width=.99\columnwidth}
\vspace{-7pt}
\end{table}

\begin{table}[t]
\caption{Comparison with Lin \etal~\cite{lin2021continual} on CIFAR100 (2 and 5 tasks, class-incremental setting). MoCoV2+ is an updated version of MoCoV2 that uses a symmetric loss. The difference between the two is $\approx$1\% at convergence~\cite{chen2021exploring}.}
\label{tab:comp-with-lin}
\vspace{-5pt}
\scriptsize
\centering
\captionsetup{type=table}
\begin{tabular}{lccccc}
\toprule
\textbf{Strategy}                       & \textbf{Method}         & \textbf{2 Tasks} & \textbf{5 Tasks} \\ 
\midrule
\multirow{2}{*}{Lin \etal\cite{lin2021continual}}      & \CC{ftcolor}SimCLR & \CC{ftcolor}55.7 & \CC{ftcolor}- \\
                             & \CC{baselinecolor}MoCoV2 
                             & \CC{baselinecolor}56.1 & \CC{baselinecolor}53.8 \\ 
\midrule
  \CC{contrcolor}  & \CC{ftcolor}SimCLR & \CC{contrcolor}\textbf{61.8} & \CC{contrcolor}\textbf{58.3}\\
       \multirow{-2}{*}{\CC{contrcolor}\name{}}                        & \CC{ftcolor}MoCoV2+ 
                             & \CC{contrcolor}\textbf{63.3} & \CC{contrcolor}\textbf{59.5} \\ 
\bottomrule
\end{tabular}
\captionsetup{width=.99\linewidth}
\vspace{-14pt}
\end{table}

\subsection{Results}\vspace{-3mm}
\noindent\textbf{Comparison with the state of the art.} In Tab.~\ref{tab:comp-with-sota} we report comparison with CL baselines and fine-tuning in composition with three SSL methods: SimCLR, Barlow Twins and BYOL. We select these three methods for the following reasons: (i) they feature different losses (InfoNCE, Cross-correlation and MSE), (ii) they exhibit different feature normalizations ($l2$, standardization and mean centering), and (iii) they use different techniques to avoid collapse (negatives, redundancy reduction, momentum encoder). The comparison is performed on class-incremental CIFAR100 with 5 tasks. Offline learning results are reported as upper bound. 

\begin{figure}[t]
    \centering
    \includegraphics[width=\columnwidth]{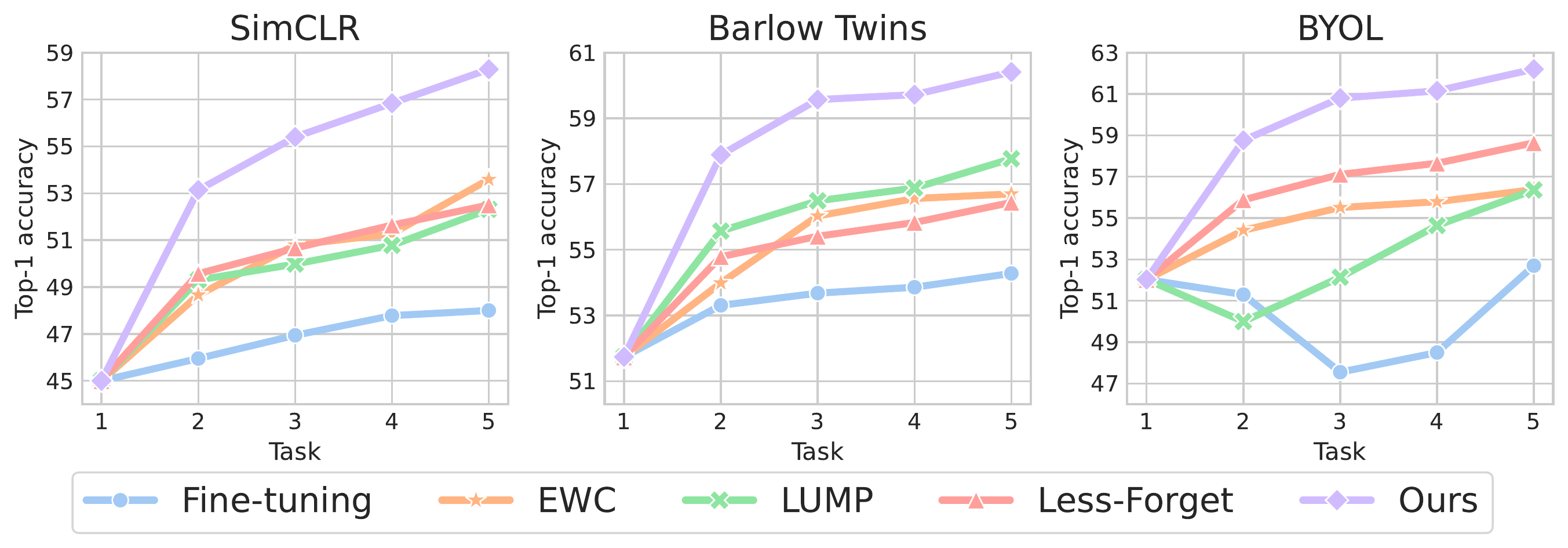}
    \vspace{-22pt}
    \caption{Evolution of top-1 linear evaluation accuracy over tasks on CIFAR100 (5 tasks, class-incremental).}
    \label{fig:lineplot}
    \vspace{-13pt}
\end{figure}

First, we notice that \name{} produces better representations than all the other strategies, outperforming them by large margins with all SSL methods in terms of top-1 accuracy. Moreover, our framework also shows better forward transfer, meaning that its features are easier to generalize to other tasks (also evident in Tab.~\ref{tab:im100-to-domainnet}). \name{} appears to reduce catastrophic forgetting with respect to fine-tuning, and is comparable to other methods. 
In general, SSL methods already have low forgetting with respect to supervised learning on CIFAR100 (see Tab.~\ref{tab:c100-in100-class-inc}) and therefore there is little margin for improvement. However, on higher resolution images (ImageNet100) \name{} actually achieves remarkable results in the mitigation of catastrophic forgetting.

Replay-based methods (ER, DER) clearly do not help against forgetting in CSSL. We found two reasons for this failure. First, in supervised CL, replay-based methods benefit from storing labels, which contain a lot of information about previous tasks and enable the retraining of the linear classifier on old classes. This is not the case in CSSL, where labels are unavailable. Second, SSL models need more training epochs to converge, which means that samples in the buffer are also replayed many more times. This causes severe overfitting on these exemplars, defeating the purpose of the replay buffer. LUMP mitigates this effect by augmenting the buffer using mixup but does not reach too far, surpassing other baselines only with Barlow Twins. EWC holds up surprisingly well, outperforming more recent methods, meaning that the importance of the weights can be calculated accurately with the self-supervised loss. Distillation methods (POD, Less-Forget) show good performance. However, they use $l2$-normalization in their loss, causing loss of information when coupled with Barlow Twins, which decreases accuracy.

\begin{table}
\caption{Linear evaluation top-1 accuracy on class-incremental CIFAR100 and ImageNet100 with 5 tasks. \name{} is compared to fine-tuning, offline and supervised learning.}
\label{tab:c100-in100-class-inc}
\vspace{-7pt}
\setlength{\tabcolsep}{4.9pt}
\scriptsize
\centering
\captionsetup{type=table}
\begin{tabular}{lccccccc}
\toprule
\multirow{2}[1]{*}{\textbf{Method}} & \multirow{2}[1]{*}{\textbf{Strategy}}  & \multicolumn{3}{c}{\textbf{CIFAR100}} & \multicolumn{3}{c}{\textbf{ImageNet100}} \\
\cmidrule(lr){3-5}\cmidrule(lr){6-8}
&& \textbf{A ($\uparrow$)} & \textbf{F ($\downarrow$)} & \textbf{T ($\uparrow$)} & \textbf{A ($\uparrow$)} & \textbf{F ($\downarrow$)} & \textbf{T ($\uparrow$)}  \\ 
\midrule
\multirow{3}[1]{*}{{\parbox{0.75cm}{Barlow\\Twins}}}      & \CC{ftcolor}Fine-tuning & \CC{ftcolor}54.3 & \CC{ftcolor}\textbf{0.4} & 39.2 & \CC{ftcolor}63.1 & \CC{ftcolor}10.7 & 44.4\\
                             & \CC{decorrcolor}\name{} 
                             &\CC{decorrcolor}\textbf{60.4} &\CC{decorrcolor}\textbf{0.4} & \CC{decorrcolor}\textbf{42.2} & \CC{decorrcolor}\textbf{68.2} & \CC{decorrcolor}\textbf{1.3} & \CC{decorrcolor}\textbf{47.9} \\ 
                             \cmidrule{2-8}
                             & \CC{offlinecolor} Offline & \CC{offlinecolor}70.9 & \CC{offlinecolor}- & \CC{offlinecolor}- & \CC{offlinecolor}80.4 & \CC{offlinecolor}- & \CC{offlinecolor}- \\
\midrule
\multirow{3}[1]{*}{SwAV}     & \CC{ftcolor}Fine-tuning & \CC{ftcolor}55.5 & \CC{ftcolor}\textbf{0.0} & \CC{ftcolor}32.8 & \CC{ftcolor}64.4 & \CC{ftcolor}4.3 & \CC{ftcolor} 42.8 \\
                             & \CC{knowcolor}\name{} 
                             & \CC{knowcolor}\textbf{57.8} & \CC{knowcolor}\textbf{0.0} & \CC{knowcolor}\textbf{34.5} & \CC{knowcolor}\textbf{66.0} & \CC{knowcolor}\textbf{0.2}  & \CC{knowcolor}\textbf{43.6}  \\
                             \cmidrule{2-8}
                             & \CC{offlinecolor} Offline & \CC{offlinecolor}64.9 & \CC{offlinecolor}-&  \CC{offlinecolor}-& \CC{offlinecolor}74.3& \CC{offlinecolor}-& \CC{offlinecolor}-\\
\midrule
\multirow{3}[1]{*}{BYOL}      & \CC{ftcolor}Fine-tuning & \CC{ftcolor}52.7 & \CC{ftcolor}0.1 & \CC{ftcolor}35.9 & \CC{ftcolor}66.0 & \CC{ftcolor}2.9 & \CC{ftcolor}43.2 \\
                             & \CC{predcolor}\name{} 
                             & \CC{predcolor}\textbf{62.2} & \CC{predcolor}\textbf{0.0} & \CC{predcolor}\textbf{42.2} & \CC{predcolor}\textbf{66.4} & \CC{predcolor}\textbf{1.1} & \CC{predcolor}\textbf{46.6}  \\
                             \cmidrule{2-8}
                             & \CC{offlinecolor} Offline & \CC{offlinecolor}70.5 & \CC{offlinecolor}-& \CC{offlinecolor}- & \CC{offlinecolor}80.3  & \CC{offlinecolor}-& \CC{offlinecolor}-\\
\midrule
\multirow{3}[1]{*}{VICReg}      & \CC{ftcolor}Fine-tuning & \CC{ftcolor}51.5 & \CC{ftcolor}0.9 & \CC{ftcolor}36.4  & \CC{ftcolor}61.3 & \CC{ftcolor}7.9 & \CC{ftcolor}42.0  \\
                             & \CC{predcolor}\name{} 
                             & \CC{predcolor}\textbf{53.6} & \CC{predcolor}\textbf{0.2}  & \CC{predcolor}\textbf{41.1} & \CC{predcolor}\textbf{64.8} & \CC{predcolor}\textbf{4.3} & \CC{predcolor}\textbf{45.3} \\
                             \cmidrule{2-8}
                             & \CC{offlinecolor} Offline & \CC{offlinecolor}68.5  & \CC{offlinecolor}- & \CC{offlinecolor}- & \CC{offlinecolor}79.4 & \CC{offlinecolor}- & \CC{offlinecolor}-   \\
\midrule
\multirow{3}[1]{*}{MoCoV2+}      & \CC{ftcolor}Fine-tuning & \CC{ftcolor}47.3 & \CC{ftcolor}0.2 & \CC{ftcolor}33.4 & \CC{ftcolor}62.0 & \CC{ftcolor}8.4 & \CC{ftcolor}41.6  \\
                             & \CC{contrcolor}\name{} 
                             & \CC{contrcolor}\textbf{59.5} & \CC{contrcolor}\textbf{0.0}  & \CC{contrcolor}\textbf{39.6} & \CC{contrcolor}\textbf{68.8} & \CC{contrcolor}\textbf{1.5}  & \CC{contrcolor}\textbf{46.8} \\
                             \cmidrule{2-8}
                             & \CC{offlinecolor} Offline & \CC{offlinecolor}69.9 & \CC{offlinecolor}- & \CC{offlinecolor}- & \CC{offlinecolor}79.3 & \CC{offlinecolor}- & \CC{offlinecolor}- \\
\midrule
\multirow{3}[1]{*}{SimCLR}      & \CC{ftcolor}Fine-tuning & \CC{ftcolor}48.9 & \CC{ftcolor}1.0 & \CC{ftcolor}33.5 & \CC{ftcolor}61.5 & \CC{ftcolor}8.1  & \CC{ftcolor}40.3 \\
                             & \CC{contrcolor}\name{} 
                             & \CC{contrcolor}\textbf{58.3} & \CC{contrcolor}\textbf{0.2} & \CC{contrcolor}\textbf{36.4} & \CC{contrcolor}\textbf{68.0} & \CC{contrcolor}\textbf{2.2} & \CC{contrcolor}\textbf{45.8}  \\ 
                             \cmidrule{2-8} 
                             & \CC{offlinecolor} Offline & \CC{offlinecolor}65.8 & \CC{offlinecolor}-& \CC{offlinecolor}- & \CC{offlinecolor}77.5 & \CC{offlinecolor}-& \CC{offlinecolor}- \\
\midrule
\multirow{2}[1]{*}{Supervised}  & \CC{ftcolor}Fine-tuning & \CC{ftcolor}54.1 & \CC{ftcolor}6.8 & \CC{ftcolor}36.5   & \CC{ftcolor}63.1 & \CC{ftcolor}5.6 & \CC{ftcolor}42.5 \\
                            \cmidrule{2-8}
                             & \CC{offlinecolor} Offline & \CC{offlinecolor}75.6 & \CC{offlinecolor}- & \CC{offlinecolor}- & \CC{offlinecolor}81.9 & \CC{offlinecolor}- & \CC{offlinecolor}-  \\
\bottomrule
\end{tabular}
\captionsetup{width=.99\columnwidth}
\vspace{-7pt}
\end{table}

Fig.~\ref{fig:lineplot} shows the evolution of top-1 linear evaluation accuracy over the whole training trajectory on class-incremental CIFAR100 with 5 tasks. \name{} outperforms the other methods, and keeps improving throughout the sequence. We found BYOL to be unstable when simply fine-tuning the model. \name{}, EWC and Less-Forget mitigate this instability completely. On the other hand, LUMP first drops slightly and then recovers. We believe this is due to some instability introduced by the mixup regularization, to which the model takes time to adapt.

In Tab.~\ref{tab:comp-with-lin} we also compare with Lin \etal~\cite{lin2021continual} on class-incremental CIFAR100. Although our method is not specifically designed for contrastive learning, it substantially outperforms Lin \etal with 2 and 5 tasks. It is worth nothing that MoCoV2+ is slightly better than MoCoV2 ($\approx$1\% difference), whereas our gains are much larger ($\approx$7\%).

\noindent\textbf{Ablation study.} We ablate the most critical design choices we adopt in \name{}: (i) distillation without swapped views, and (ii) the presence of a prediction head $g$. These results are reported in Tab.~\ref{tab:ablation}. Our full framework clearly outperforms its variants with swapped views and without predictor. This validates our hypothesis that a predictor to map new features to the old feature space is crucial. The result that swapping views does not help is likely due to the frozen encoder not being invariant to the current task.

\noindent\textbf{Class-incremental.} In Tab.~\ref{tab:c100-in100-class-inc} we report a study of CSSL with 6 SSL methods in composition with the \name{} framework on class-incremental CIFAR100 and ImageNet100. Fine-tuning and Offline SSL results are reported as lower and upper bounds. The accuracy of supervised learning is also reported. \name{} always improves with respect to fine-tuning. In particular, our framework produces higher forward transfer and lower forgetting, especially on ImageNet100, where methods tend to forget more. Notably, \name{} outperforms supervised fine-tuning, except when coupled with VICReg on CIFAR100. On average, SSL methods trained continually with \name{} improve by 6.8\% on CIFAR100 and 4\% on ImageNet100.
\begin{table}[t]
\caption{Training 5 times longer on 1/5 of the data vs. training continually w/ and w/o \name{}  on ImageNet100 (5 tasks, class- and data-incremental). \textbf{Bold} is best, \underline{underlined} is second best.}
\label{tab:longer-vs-continual}
\vspace{-8pt}
\scriptsize
\centering
\captionsetup{type=table}
\begin{tabular}{lcccccc}
\toprule
\textbf{Setting} &\textbf{Method}                       & \textbf{Fine-tune}         & \textbf{Offline 1/5} & \textbf{\name{}} \\ 
\midrule
\multirow{3}{*}{Class-inc.} & SimCLR & 61.5 & \underline{63.1} & \CC{contrcolor}\textbf{68.0} \\
& Barlow Twins & 63.1 & \underline{63.5} & \CC{decorrcolor}\textbf{68.2} \\
& BYOL & \underline{66.0} & 60.6 & \CC{predcolor}\textbf{66.4} \\
\midrule
\multirow{3}{*}{Data-inc.} & SimCLR & \underline{68.9} & 67.2 & \CC{contrcolor}\textbf{72.1} \\
& Barlow Twins & \underline{71.3} & 70.2 & \CC{decorrcolor}\textbf{74.9} \\
& BYOL & \textbf{74.0} & 66.7 & \CC{predcolor}\underline{73.3} \\
\bottomrule
\end{tabular}
\captionsetup{width=.99\linewidth}
\vspace{-8pt}
\end{table}
\begin{table}[t]
\caption{Ablation study of design choices in \name{}.}
\label{tab:ablation}
\vspace{-8pt}
\scriptsize
\centering
\captionsetup{type=table}
\begin{tabular}{lcccc}
\toprule
\textbf{Strategy}                       & \textbf{Method}         & \textbf{Swap} & \textbf{No pred.} & \textbf{Ours}\\ 
\midrule
\multirow{3}{*}{\name{}}      & SimCLR & \CC{contrcolor}49.3 & \CC{contrcolor}52.6 &  \CC{contrcolor}\textbf{58.3} \\
                             & Barlow Twins  & \CC{decorrcolor}57.4 & \CC{decorrcolor}57.3 & \CC{decorrcolor}\textbf{60.4}  \\ 
                             & BYOL  & \CC{predcolor}52.0 & \CC{predcolor}58.6 & \CC{predcolor}\textbf{62.2} \\ 
\bottomrule
\end{tabular}
\captionsetup{width=.99\linewidth}
\vspace{-12pt}
\end{table}

\noindent\textbf{Data-incremental.}
Tab.~\ref{tab:domain-data-incremental} presents results for linear evaluation top-1 accuracy on ImageNet100 with 5 tasks in a data-incremental scenario. While no SSL method is better than supervised fine-tuning, Barlow Twins coupled with \name{} is competitive. \name{} improves performance in all cases by 2\% on average, except for BYOL. This is likely due to the fact that in the data-incremental scenario remembering past knowledge is less important than in other scenarios, and BYOL already has a momentum encoder that provides some information about the past. This hypothesis is validated by the fact that MoCoV2+ (that uses a momentum encoder) improves less than SimCLR when coupled with \name{}. We believe that, by tuning the EMA schedule, improvement could also be achieved for BYOL. In addition, BYOL already shows impressive performance with fine-tuning, outperforming all the other methods by more than 2\%. Interestingly, SwAV comes closest to its offline upper bound, with only a 3\% decrease in performance when coupled with \name{}.

\begin{table}[t]
\caption{Linear evaluation accuracy on ImageNet100 (5 tasks, data-incremental) and DomainNet (6 tasks, domain-incremental).}
\label{tab:domain-data-incremental}
\vspace{-8pt}
\setlength{\tabcolsep}{4pt}
\scriptsize
\centering
\captionsetup{type=table}
\begin{tabular}{lccc}
\toprule
\textbf{Method} & \textbf{Strategy} & \parbox{1.5cm}{\centering\textbf{ImageNet100}\\\textbf{(Data-inc.)}} & \parbox{1.5cm}{\centering\textbf{DomainNet}\\\textbf{(Domain-inc.)}} \\ 
\midrule
\multirow{3}[1]{*}{{\parbox{1cm}{Barlow\\Twins}}}      & \CC{ftcolor}Fine-tuning & \CC{ftcolor}71.3 & 50.3 \\
                             & \CC{decorrcolor}\name{} 
                             & \CC{decorrcolor}\textbf{74.9} & \CC{decorrcolor}\textbf{55.5} \\ 
                             \cmidrule{2-4}
                             & \CC{offlinecolor} Offline & \CC{offlinecolor}80.4 & \CC{offlinecolor}57.2  \\
\midrule
\multirow{3}[2]{*}{SwAV}      & \CC{ftcolor}Fine-tuning & \CC{ftcolor}70.8 & 49.6 \\
                             & \CC{knowcolor}Knowledge 
                             & \CC{knowcolor}\textbf{71.3} & \CC{knowcolor}\textbf{54.3} \\ 
                             \cmidrule{2-4}
                             & \CC{offlinecolor} Offline & \CC{offlinecolor}74.3 & \CC{offlinecolor}54.6 \\
\midrule
\multirow{3}[2]{*}{BYOL}      & \CC{ftcolor}Fine-tuning & \CC{ftcolor}\textbf{74.0} & 50.6 \\
                             & \CC{predcolor}\name{} 
                             & \CC{predcolor}73.3 & \CC{predcolor}\textbf{55.1}\\ \cmidrule{2-4}
                             & \CC{offlinecolor} Offline & \CC{offlinecolor}80.3 & \CC{offlinecolor}56.6 \\
\midrule
\multirow{3}[2]{*}{VICReg}      & \CC{ftcolor}Fine-tuning & \CC{ftcolor}70.2 & 49.3  \\
                             & \CC{predcolor}\name{} 
                             & \CC{predcolor}\textbf{72.3} & \CC{predcolor}\textbf{52.9} \\ 
                             \cmidrule{2-4}
                             & \CC{offlinecolor} Offline & \CC{offlinecolor}79.4 & \CC{offlinecolor}56.7 \\
\midrule
\multirow{3}[2]{*}{MoCoV2+}      & \CC{ftcolor}Fine-tuning & \CC{ftcolor}69.5 & 43.2 \\
                             & \CC{contrcolor}\name{} 
                             & \CC{contrcolor}\textbf{71.9} & \CC{contrcolor}\textbf{46.7} \\ 
                             \cmidrule{2-4}
                             & \CC{offlinecolor} Offline & \CC{offlinecolor}78.2 &\CC{offlinecolor}53.7 \\
\midrule
\multirow{3}[2]{*}{SimCLR}      & \CC{ftcolor}Fine-tuning & \CC{ftcolor}68.9 & 45.1  \\
                             & \CC{contrcolor}\name{} 
                             & \CC{contrcolor}\textbf{72.1} & \CC{contrcolor}\textbf{50.0} \\
                             \cmidrule{2-4}
                             & \CC{offlinecolor} Offline & \CC{offlinecolor}77.5 & \CC{offlinecolor}52.6 \\
\midrule
\multirow{2}[1]{*}{Supervised}   & \CC{ftcolor}Fine-tuning & \CC{ftcolor}75.9 & 55.9 \\
                             \cmidrule{2-4}
                             & \CC{offlinecolor} Offline & \CC{offlinecolor}81.9 & 66.4  \\
\bottomrule
\end{tabular}
\captionsetup{width=.99\linewidth}
\vspace{-8pt}
\end{table}
\noindent\textbf{Domain-incremental.} We also examine the capability of \name{} to learn continually when the domain from which the data is drawn changes. Tab.~\ref{tab:domain-data-incremental} shows the average top-1 accuracy of a linear classifier trained on top of the frozen feature extractor on all domains separately (domain-aware). Domain-agnostic evaluation and results for each domain are presented in the supplementary material. Again, \name{} improves every method by 4.4\% on average, showing that our distillation strategy is robust to domain shift, and although the data distribution is really different, information transfer is still performed. Interestingly, most of the methods, when trained with \name{} get very close to their offline accuracy.

\noindent\textbf{Long training vs continual training.} We also analyze the following question: is it worth training continually or is it better to train for longer on a small dataset? This depends on two factors: (i) the SSL method, and (ii) the CSSL setting. For SimCLR and Barlow Twins in the class-incremental setting it seems to be better to train offline on 1/5 of the classes instead of training continually with 5 tasks. In this setting, offline BYOL seems to suffer from instability, ending up lower than fine-tuning. On the other hand, on the data-incremental setting, fine-tuning outperforms longer training, especially for BYOL, which also outperforms \name{} (as explained previously). Apart from this exception, \name{} always produces better representations than other strategies, making it the go-to option.

\noindent\textbf{Downstream and semi-supervised.}
In Tab.~\ref{tab:im100-to-domainnet}, we present the downstream performance of \name{} compared with fine-tuning when trained on ImageNet100 and evaluated on DomainNet (Real). Barlow Twins, SwAV and BYOL show higher performance than the supervised model, even when considering a fine-tuning strategy. This is probably due to the fact that SSL methods tend to learn more general features than their supervised counterparts. \name{} improves performance on all the SSL methods, making them surpass the supervised baseline. Lastly, when compared with fine-tuning, \name{} improves the performance of SSL methods by 3.4\% on average.
Tab.~\ref{tab:imagenet100-semisup} contains the top-1 accuracy on ImageNet100 when training a linear classifier on a frozen backbone with limited amount of labels (10\% and 1\%). First, we can observe that no SSL method with fine-tune surpasses the performance of supervised learning. When using \name{}, MoCoV2+ outperforms supervised with 10\% labels and, in general, Barlow Twins and MoCoV2+ work best in both semi-supervised settings.
\name{} improves all SSL methods when compared with fine-tuning.

\begin{table}[t]
\caption{Downstream performance with different SSL methods trained on Imagenet-100 and evaluated on DomainNet (Real).}
\label{tab:im100-to-domainnet}
\vspace{-8pt}
\scriptsize
\centering
\captionsetup{type=table}
\resizebox{.47\textwidth}{!}{
\begin{tabular}{lcccccc|c}
\toprule
\parbox[c]{0.8cm}{\textbf{Strategy}} & \parbox[c]{0.8cm}{\centering \textbf{Barlow\\Twins}} & \parbox[c]{0.8cm}{\centering\textbf{SwAV}}  & \parbox[c]{0.8cm}{\centering\textbf{BYOL}}  & \parbox[c]{0.8cm}{\centering\textbf{VICReg}}  & \parbox[c]{0.8cm}{\centering\textbf{MoCoV2+}}  & \parbox[c]{0.8cm}{\centering\textbf{SimCLR}}  & \textbf{Supervised} \\ 
\midrule
Fine-tune & 56.2 & 55.9 & 55.0 & 54.0 &52.4& 51.6 & \multirow{2}{*}{54.3} \\
\name{} &\CC{decorrcolor} \textbf{60.3} &\CC{knowcolor}\textbf{56.9}&\CC{predcolor} \textbf{56.9} &\CC{predcolor}\textbf{56.3}&\CC{contrcolor}\textbf{58.7}&\CC{contrcolor}\textbf{56.5}&\\
\bottomrule
\end{tabular}
}
\vspace{-5pt}
\end{table}

\begin{table}[t]
\caption{Top-1 linear accuracy on Imagenet-100 with different SSL methods, semi-supervised setting with 10\% and 1\% of labels.}
\label{tab:imagenet100-semisup}
\vspace{-7pt}
\centering
\captionsetup{type=table}
\resizebox{0.47\textwidth}{!}{
\begin{tabular}{cccccccc|c}
\toprule
\textbf{Percentage} & \textbf{Strategy} & \parbox[c]{1.2cm}{\small\centering \textbf{Barlow\\Twins}} & \parbox[c]{1.2cm}{\small\centering\textbf{SwAV}}  & \parbox[c]{1.2cm}{\small\centering\textbf{BYOL}}  & \parbox[c]{1.2cm}{\small\centering\textbf{VICReg}}  & \parbox[c]{1.2cm}{\small\centering\textbf{MoCoV2+}}  & \parbox[c]{1.2cm}{\small\centering\textbf{SimCLR}}  & \textbf{Supervised} \\ 
\midrule
\multirow{2}{*}{10\%} & Fine-tune & 56.6 & 57.6 & 55.7 & 53.6 & 54.9 & 52.5 & \multirow{2}{*}{60.8} \\
& \name{} &\CC{decorrcolor} \textbf{60.3} &\CC{knowcolor} \textbf{58.2} &\CC{predcolor} \textbf{56.5} &\CC{predcolor} \textbf{56.5} &\CC{contrcolor} \textbf{61.7} &\CC{contrcolor} \textbf{58.9} &\\
\midrule
\multirow{2}{*}{1\%} & Fine-tune & 42.6 & 42.5 & 42.3 & 40.4 &40.9 & 39.7 & \multirow{2}{*}{48.1} \\
& \name{} &\CC{decorrcolor}\textbf{47.0} &\CC{knowcolor}\textbf{43.1} &\CC{predcolor}\textbf{43.4} &\CC{predcolor}\textbf{43.2} &\CC{contrcolor}\textbf{47.8} &\CC{contrcolor}\textbf{46.8}&\\
\bottomrule
\end{tabular}
}
\vspace{-8pt}
\end{table}

\vspace{-7pt}
\section{Conclusion}
\vspace{-8pt}
In this work, we study Continual Self-Supervised Learning (CSSL), the problem of learning a set of tasks without labels continually. We make two important contributions for the SSL and CL communities: (i) we present \name{}, a simple and effective framework for CSSL that shows how SSL methods and losses can be seamlessly reused to learn continually, and (ii) we perform a comprehensive analysis of CSSL, leading to the emergence of  interesting properties of SSL methods.

\noindent\textbf{Limitations.} Although \name{} shows exciting performance, it has some limitations. First, it is applicable in settings where task boundaries are provided. Second, our framework increases the amount of computational resources needed for training by roughly 30\%, both in terms of memory and time. Finally, \name{} does not perform clustering, meaning that it is unable to directly learn a mapping from data to latent classes, and thus needs either a linear classifier trained with supervision, or some clustering algorithm.

\noindent\textbf{Broader impact.} The capabilities of supervised CL agents are bounded by the need for human-produced annotations. CSSL models can potentially improve without the need for human supervision. This facilitates the creation of powerful AIs that may be used for malicious purposes such as discrimination and surveillance. Also, since in CSSL the data is supposed to come from a non-curated stream, the model may be affected by biases in the data. This is problematic because biases are then be transferred to downstream tasks.

\small{\noindent\textbf{Acknowledgements.} This work was supported by the European Institute of Innovation \& Technology (EIT) and the H2020 EU project SPRING, funded by the European Commission under GA 871245. It was carried out under the ``Vision and Learning joint Laboratory" between FBK and UNITN. Karteek Alahari was funded by the ANR grant AVENUE (ANR-18-CE23-0011). Julien Mairal was funded by the ERC grant number 714381 (SOLARIS project) and by ANR 3IA MIAI@Grenoble Alpes (ANR-19-P3IA-0003). Xavier Alameda-Pineda was funded by the ARN grant ML3RI (ANR-19-CE33-0008-01). This project was granted access to the HPC resources of IDRIS under the allocation 2021-[AD011013084] made by GENCI.}

\renewcommand\thesection{\Alph{section}}
\renewcommand\thefigure{\Alph{figure}}
\renewcommand\thetable{\Alph{table}}
\setcounter{section}{0}
\setcounter{figure}{0}
\setcounter{table}{0}

\section*{Appendix}

\section{PyTorch-like pseudo-code}
We provide a PyTorch-like pseudo-code of our method. As you can see, \name{} is simple to implement and does not add much complexity to the base SSL method. In this snippet, the losses are made symmetric by summing the two contributions. In some cases, the two losses are averaged instead. In \name{}, we symmetrize in the same way as the base SSL method we are considering.
\begin{algorithm}[h]

   \caption{PyTorch-like pseudo-code for \name{}.}
   \label{algo:DINO}
    \definecolor{codeblue}{rgb}{0.25,0.5,0.5}
    \definecolor{codekw}{rgb}{0.85, 0.18, 0.50}
    \lstset{
      basicstyle=\fontsize{7.2pt}{7.2pt}\ttfamily,
      commentstyle=\fontsize{7.2pt}{7.2pt}\color{codeblue},
      keywordstyle=\fontsize{7.2pt}{7.2pt}\color{codekw},
    }
\begin{lstlisting}[language=python]
# aug: stochastic image augmentation
# f: backbone and projector
# frozen_f: frozen backbone and projector
# g: CaSSLe's predictor
# loss_fn: any SSL loss in Tab. 1 (main paper)

# PyTorchLightning handles loading and optimization
def training_step(x):

    # correlated views
    x1, x2 = aug(x), aug(x)

    # forward backbone and projector
    z1, z2 = f(x1), f(x2)

    # optionally forward predictor...

    # compute SSL loss (symmetric)
    ssl_loss = loss_fn(z1, z2) \\
             + loss_fn(z2, z1)

    # forward frozen backbone and projector
    z1_bar, z2_bar = frozen_f(x1), frozen_f(x2)

    # compute distillation loss (symmetric)
    distill_loss = loss_fn(g(z1), z1_bar) \\
                 + loss_fn(g(z2), z2_bar)
    
    # no hyperparameter for loss weighting
    return ssl_loss + distill_loss
\end{lstlisting}
\end{algorithm}

\section{Derivation of distillation losses}
In this section, we derive distillation losses from the SSL losses in Tab. 1 of the main paper, starting from the definition of our distillation loss:
\begin{equation}
    \mathcal{L}_D(\zvect, \bar{\zvect}) = \mathcal{L}_{SSL} (g(\zvect), \bar{\zvect}),
\end{equation}
where $z$ and $\bar{z}$ are the representations of the current and frozen encoder, and $g$ is \name{}'s predictor network implemented as a two layer MLP with 2048 hidden neurons and ReLU activation.
\paragraph{Contrastive based.} Our distillation loss based on contrastive learning is implemented as follows:
\begin{equation}
    \mathcal{L}(z_i, \bar{z}_i) = -\log \frac{\exp \left(\operatorname{sim}\left(\zvect_i, \bar{\zvect}_i\right) / \tau\right)}{\sum_{\zvect_j \in \bar{\eta}(i)} \exp \left(\operatorname{sim}\left(\zvect_i, \zvect_{j}\right) / \tau\right)},
\end{equation}
where $\bar{\eta}(i)$ is the set of negatives for the sample with index $i$ in the batch. Note that the negatives are drawn both from the predicted and frozen features.
\paragraph{MSE based.} This distillation loss is simply the MSE between the predicted features and the frozen features:
\begin{equation}
    \mathcal{L}(z, \bar{z}) = -||g(\zvect) -  \bar{\zvect}||^2_2 .
\end{equation}
It can be implemented with the cosine similarity as stated in the main manuscript.
\paragraph{Cross-entropy based.} The cross-entropy loss, when used for distillation in an unsupervised setting, makes sure that the current encoder is able to assign samples to the frozen centroids (or prototypes) consistently with the frozen encoder:
\begin{equation}
    \mathcal{L}(z, \bar{z}) = -\sum_{d} \bar{\avect}_{d} \log \frac{\exp \left(\operatorname{sim}\left( g(\zvect), \cvect^{t-1}_{d}\right) / \tau\right)}{\sum_k \exp \left(\operatorname{sim}\left(g(\zvect), \cvect^{t-1}_k\right) / \tau\right)} 
\end{equation}
where:
\begin{equation}
    \bar{\avect}= \frac{\exp \left(\operatorname{sim}\left( \bar{\zvect}, \cvect^{t-1}_{d}\right) / \tau\right)}{\sum_k \exp \left(\operatorname{sim}\left(\bar{\zvect}, \cvect^{t-1}_k\right) / \tau\right)},
\end{equation}
and the set of frozen prototypes is denoted as follows: $\mathbf{C}^{t-1} = \left\{\mathbf{c}^{t-1}_{1}, \ldots, \mathbf{c}^{t-1}_{K}\right\}$.
 
\paragraph{Cross-correlation based.} We consider Barlow Twins'~ \cite{zbontar2021barlow} implementation of this objective. For VICReg~\cite{bardes2021vicreg} we only consider the invariance term. As a distillation loss, the cross-correlation matrix is computed with the predicted and frozen features:

\begin{equation}
    \mathcal{L}(z, \bar{z}) = \sum_{u}\left(1-\bar{\mathcal{C}}_{u v}\right)^{2}+\lambda \sum_{u} \sum_{v \neq u} \bar{\mathcal{C}}_{u v}^{2} ,
\end{equation}
where:
\begin{equation}
    \bar{\mathcal{C}}_{u v} = \frac{\sum_{i} g(\zvect_{i, u}) \bar{\zvect}_{i, v}}{\sqrt{\sum_{i}g\left(\zvect_{i, u}\right)^{2}}. \sqrt{\sum_{i}\left(\bar{\zvect}_{i, v}\right)^{2}}}.
\end{equation} 

\section{Further discussion and implementation details of the baselines}
\paragraph{Selection.} When evaluating our framework, we try to compare with as many existing related methods as possible. However, given that SSL models are computationally intensive, it was not possible to run all baselines and methods in all the CL settings we considered. As mentioned in the main manuscript, we choose eight baselines (seven related methods + fine-tuning) belonging to three CL macro-categories, and test them on CIFAR100 (class-incremental) in combination with three SSL methods. The selection was based on the ease of adaptation to CSSL and the similarity to our framework.

\begin{table*}[ht]
\centering
\scriptsize
\captionsetup{type=table}
\captionsetup{width=.99\linewidth}
\caption{Linear evaluation top-1 accuracy on DomainNet (6 tasks, domain-incremental setting) w/ and w/o \name{}. The sequence of tasks is Real$\rightarrow$Quickdraw$\rightarrow$Painting$\rightarrow$Sketch$\rightarrow$Infograph$\rightarrow$Clipart. ``Aw.'' stands for task-aware, ``Ag,'' for task-agnostic.}
\label{tab:domain-incremental-task-agnostic}
\vspace{-2px}
\begin{tabular}{lccccccccccccccc}
\toprule
\multirow{2}[1]{*}{\textbf{Method}} & \multirow{2}[1]{*}{\textbf{Strategy}} & \multicolumn{2}{c}{\textbf{Real}} & \multicolumn{2}{c}{\textbf{Quickdraw}} & \multicolumn{2}{c}{\textbf{Painting}} & \multicolumn{2}{c}{\textbf{Sketch}} & \multicolumn{2}{c}{\textbf{Infograph}} & \multicolumn{2}{c}{\textbf{Clipart}} & \multicolumn{2}{c}{\textbf{Avg.}} \\ 
\cmidrule(lr){3-4} \cmidrule(lr){5-6} \cmidrule(lr){7-8} \cmidrule(lr){9-10} \cmidrule(lr){11-12} \cmidrule(lr){13-14} \cmidrule(lr){15-16}
&& Aw. & Ag. & Aw. & Ag. & Aw. & Ag. & Aw. & Ag. & Aw. & Ag. & Aw. & Ag. & Aw. & Ag.   \\
\midrule
\multirow{3}[2]{*}{{\parbox{1.5cm}{Barlow Twins}}} & \CC{ftcolor}Finetuning & \CC{ftcolor}56.3 & \CC{ftcolor}50.9 & \CC{ftcolor}54.1 & \CC{ftcolor}45.8 & \CC{ftcolor}42.7 & \CC{ftcolor}35.9 & \CC{ftcolor}49.0 & \CC{ftcolor}41.9 & \CC{ftcolor}22.0 & \CC{ftcolor}17.4 & \CC{ftcolor}59.0 & \CC{ftcolor}52.5 & \CC{ftcolor}50.3 & \CC{ftcolor}43.7 \\
                             & \CC{decorrcolor}\name{} 
                             & \CC{decorrcolor}\textbf{62.7} & \CC{decorrcolor}\textbf{57.1} & \CC{decorrcolor}\textbf{59.1} & \CC{decorrcolor}\textbf{50.6} & \CC{decorrcolor}\textbf{49.2} & \CC{decorrcolor}\textbf{42.1} & \CC{decorrcolor}\textbf{53.8} & \CC{decorrcolor}\textbf{47.7} & \CC{decorrcolor}\textbf{25.5} & \CC{decorrcolor}\textbf{20.6} & \CC{decorrcolor}\textbf{61.9} & \CC{decorrcolor}\textbf{55.6} & \CC{decorrcolor}\textbf{55.5} & \CC{decorrcolor}\textbf{48.9} \\ 
                             \cmidrule{2-16}
                             & \CC{offlinecolor} Offline & \CC{offlinecolor}67.1 & \CC{offlinecolor}63.0 & \CC{offlinecolor}60.3 & \CC{offlinecolor}53.9 & \CC{offlinecolor}52.4 & \CC{offlinecolor}46.3 & \CC{offlinecolor}51.9 & \CC{offlinecolor}46.9 & \CC{offlinecolor}25.9 & \CC{offlinecolor}21.0 & \CC{offlinecolor}58.8 & \CC{offlinecolor}52.6 & \CC{offlinecolor}57.2 & \CC{offlinecolor}51.8 \\
\midrule
\multirow{3}[2]{*}{SwAV}      & \CC{ftcolor}Finetuning & \CC{ftcolor}57.7 & \CC{ftcolor}52.3 & \CC{ftcolor}53.2 & \CC{ftcolor}43.5 & \CC{ftcolor}43.0 & \CC{ftcolor}35.9 & \CC{ftcolor}46.1 & \CC{ftcolor}39.0 & \CC{ftcolor}21.6 & \CC{ftcolor}16.5 & \CC{ftcolor}53.4 & \CC{ftcolor}46.6 & \CC{ftcolor}49.6 & \CC{ftcolor}42.5 \\
                             & \CC{knowcolor}\name{} 
                             & \CC{knowcolor}\textbf{62.8} & \CC{knowcolor}\textbf{57.8} & \CC{knowcolor}\textbf{59.5} & \CC{knowcolor}\textbf{50.2} & \CC{knowcolor}\textbf{47.5} & \CC{knowcolor}\textbf{41.2} & \CC{knowcolor}\textbf{49.5} & \CC{knowcolor}\textbf{42.5} & \CC{knowcolor}\textbf{22.5} & \CC{knowcolor}\textbf{17.9} & \CC{knowcolor}\textbf{56.5} & \CC{knowcolor}\textbf{49.6} & \CC{knowcolor}\textbf{54.3} & \CC{knowcolor}\textbf{47.5} \\
                             \cmidrule{2-16}
                             & \CC{offlinecolor} Offline  & \CC{offlinecolor}64.1  & \CC{offlinecolor}59.5 & \CC{offlinecolor}60.6 & \CC{offlinecolor}53.6 & \CC{offlinecolor}47.6 & \CC{offlinecolor}42.9 & \CC{offlinecolor}47.7 & \CC{offlinecolor}42.1 & \CC{offlinecolor}23.3 & \CC{offlinecolor}18.9 & \CC{offlinecolor}53.6 & \CC{offlinecolor}47.3 & \CC{offlinecolor}54.6 & \CC{offlinecolor}49.1 \\
\midrule
\multirow{3}[2]{*}{BYOL}      & \CC{ftcolor}Finetuning & \CC{ftcolor}58.7 & \CC{ftcolor}53.2& \CC{ftcolor}51.7 & \CC{ftcolor}41.6 & \CC{ftcolor}44.0 & \CC{ftcolor}37.4 & \CC{ftcolor}49.6 & \CC{ftcolor}43.9 & \CC{ftcolor}23.5 & \CC{ftcolor}19.0 & \CC{ftcolor}58.6 & \CC{ftcolor}53.5 & \CC{ftcolor}50.6 & \CC{ftcolor}43.8 \\
                             & \CC{predcolor}\name{} 
                             & \CC{predcolor}\textbf{63.7} & \CC{predcolor}\textbf{60.5} & \CC{predcolor}\textbf{59.3} & \CC{predcolor}\textbf{50.9} & \CC{predcolor}\textbf{48.6} & \CC{predcolor}\textbf{44.1} & \CC{predcolor}\textbf{50.4} & \CC{predcolor}\textbf{45.2} & \CC{predcolor}\textbf{24.1} & \CC{predcolor}\textbf{19.4} & \CC{predcolor}\textbf{59.0} & \CC{predcolor}\textbf{54.4} & \CC{predcolor}\textbf{55.1} & \CC{predcolor}\textbf{49.7} \\
                             \cmidrule{2-16}
                             & \CC{offlinecolor} Offline & \CC{offlinecolor}67.2 & \CC{offlinecolor}64.0 & \CC{offlinecolor}60.2 & \CC{offlinecolor}53.3 & \CC{offlinecolor}51.5 & \CC{offlinecolor}47.3 & \CC{offlinecolor}50.4 &  \CC{offlinecolor}46.2 & \CC{offlinecolor}24.5 & \CC{offlinecolor}20.8 & \CC{offlinecolor}57.0 & \CC{offlinecolor}51.5 & \CC{offlinecolor}56.6  & \CC{offlinecolor}51.9  \\
\midrule
\multirow{3}[2]{*}{VICReg}      & \CC{ftcolor}Finetuning & \CC{ftcolor}54.7 & \CC{ftcolor}49.6 & \CC{ftcolor}53.0 & \CC{ftcolor}44.9 & \CC{ftcolor}42.1 & \CC{ftcolor}34.7 & \CC{ftcolor}49.0 & \CC{ftcolor}41.9 & \CC{ftcolor}21.1 & \CC{ftcolor}16.4 & \CC{ftcolor}58.5 & \CC{ftcolor}52.6 & \CC{ftcolor}49.3 & \CC{ftcolor}42.8  \\
                             & \CC{predcolor}\name{} 
                             & \CC{predcolor}\textbf{59.0} & \CC{predcolor}\textbf{53.2} & \CC{predcolor}\textbf{56.4} & \CC{predcolor}\textbf{47.8} & \CC{predcolor}\textbf{46.0} & \CC{predcolor}\textbf{38.9} & \CC{predcolor}\textbf{52.3} & \CC{predcolor}\textbf{45.6} & \CC{predcolor}\textbf{23.9} & \CC{predcolor}\textbf{18.5} & \CC{predcolor}\textbf{60.9} & \CC{predcolor}\textbf{55.3} & \CC{predcolor}\textbf{52.9} & \CC{predcolor}\textbf{46.1} \\ 
                             \cmidrule{2-16}
                             & \CC{offlinecolor} Offline & \CC{offlinecolor}66.4 &  \CC{offlinecolor}62.7 & \CC{offlinecolor}59.2 & \CC{offlinecolor}53.5 & \CC{offlinecolor}52.4 & \CC{offlinecolor}47.2 & \CC{offlinecolor}53.2 & \CC{offlinecolor}48.1 & \CC{offlinecolor}25.3 & \CC{offlinecolor}20.7 & \CC{offlinecolor}58.3 & \CC{offlinecolor}53.2 & \CC{offlinecolor}56.7 & \CC{offlinecolor}51.9 \\
\midrule
\multirow{3}[2]{*}{SimCLR}      & \CC{ftcolor}Finetuning & \CC{ftcolor}52.5 & \CC{ftcolor}47.6 & \CC{ftcolor}48.2 & \CC{ftcolor}38.1 & \CC{ftcolor}37.5 & \CC{ftcolor}31.7 & \CC{ftcolor}42.8 & \CC{ftcolor}35.7 & \CC{ftcolor}18.8 & \CC{ftcolor}14.4 & \CC{ftcolor}50.9 & \CC{ftcolor}46.8 & \CC{ftcolor}45.1 & \CC{ftcolor}38.4 \\
                             & \CC{contrcolor}\name{} 
                             & \CC{contrcolor}\textbf{58.4} & \CC{contrcolor}\textbf{43.4} & \CC{contrcolor}\textbf{54.2} & \CC{contrcolor}\textbf{44.7} & \CC{contrcolor}\textbf{43.9} & \CC{contrcolor}\textbf{37.7} & \CC{contrcolor}\textbf{47.6} & \CC{contrcolor}\textbf{41.9} & \CC{contrcolor}\textbf{22.0} & \CC{contrcolor}\textbf{17.8} & \CC{contrcolor}\textbf{54.9} & \CC{contrcolor}\textbf{50.5} & \CC{contrcolor}\textbf{50.0} & \CC{contrcolor}\textbf{44.2} \\ 
                             \cmidrule{2-16}
                             & \CC{offlinecolor} Offline & \CC{offlinecolor}62.1 & \CC{offlinecolor}59.5& \CC{offlinecolor}58.3 & \CC{offlinecolor}52.9 & \CC{offlinecolor}46.1 & \CC{offlinecolor}42.5 & \CC{offlinecolor}45.6 & \CC{offlinecolor}41.3 & \CC{offlinecolor}22.1 & \CC{offlinecolor}18.8 & \CC{offlinecolor}51.0 & \CC{offlinecolor}45.9 & \CC{offlinecolor}52.6 & \CC{offlinecolor}48.6 \\
\midrule
\multirow{3}[2]{*}{MoCoV2+}      & \CC{ftcolor}Finetuning & \CC{ftcolor}50.9 & \CC{ftcolor}45.5 & \CC{ftcolor}45.8 & \CC{ftcolor}37.5 & \CC{ftcolor}36.0 & \CC{ftcolor}29.3 & \CC{ftcolor}39.5 & \CC{ftcolor}32.1 & \CC{ftcolor}17.9 & \CC{ftcolor}13.5 & \CC{ftcolor}50.3 & \CC{ftcolor}\textbf{44.5} & \CC{ftcolor}43.2 & \CC{ftcolor}36.7 \\
                             & \CC{contrcolor}\name{} 
                             & \CC{contrcolor}\textbf{56.0} & \CC{contrcolor}\textbf{50.3}  & \CC{contrcolor}\textbf{48.7} & \CC{contrcolor}\textbf{40.0} & \CC{contrcolor}\textbf{40.4} & \CC{contrcolor}\textbf{33.6} & \CC{contrcolor}\textbf{42.0} & \CC{contrcolor}\textbf{35.0} & \CC{contrcolor}\textbf{19.9} & \CC{contrcolor}\textbf{15.2} & \CC{contrcolor}\textbf{51.7} & \CC{contrcolor}\textbf{44.5} & \CC{contrcolor}\textbf{46.7} & \CC{contrcolor}\textbf{38.8} \\ 
                             \cmidrule{2-16}
                             & \CC{offlinecolor} Offline & \CC{offlinecolor}65.2 & \CC{offlinecolor}61.3 & \CC{offlinecolor}57.9 & \CC{offlinecolor}51.3 & \CC{offlinecolor}48.7 & \CC{offlinecolor}43.1 & \CC{offlinecolor}44.7 & \CC{offlinecolor}39.1 & \CC{offlinecolor}23.4 & \CC{offlinecolor}19.0 & \CC{offlinecolor}51.3 & \CC{offlinecolor}44.8 & \CC{offlinecolor}53.7 & \CC{offlinecolor}48.4 \\
\midrule
\multirow{3}[2]{*}{{\parbox{1.5cm}{Supervised Contrastive}}}      & \CC{ftcolor}Finetuning & \CC{ftcolor}57.7 & \CC{ftcolor}52.6 & \CC{ftcolor}55.3 & \CC{ftcolor}45.5 & \CC{ftcolor}44.9 & \CC{ftcolor}38.0 & \CC{ftcolor}51.7 & \CC{ftcolor}45.0 & \CC{ftcolor}22.6 & \CC{ftcolor}18.3 & \CC{ftcolor}64.0 & \CC{ftcolor}60.0 & \CC{ftcolor}52.1 & \CC{ftcolor}45.4 \\
                             & \CC{contrcolor}\name{} 
                             & \CC{contrcolor}\textbf{63.4} & \CC{contrcolor}\textbf{58.8} & \CC{contrcolor}\textbf{59.7} & \CC{contrcolor}\textbf{51.3} & \CC{contrcolor}\textbf{50.1} & \CC{contrcolor}\textbf{44.7} & \CC{contrcolor}\textbf{55.9} & \CC{contrcolor}\textbf{50.3} & \CC{contrcolor}\textbf{26.9} & \CC{contrcolor}\textbf{22.4} & \CC{contrcolor}\textbf{65.0} & \CC{contrcolor}\textbf{61.3} & \CC{contrcolor}\textbf{56.7} & \CC{contrcolor}\textbf{50.9} \\
                             \cmidrule{2-16}
                             & \CC{offlinecolor} Offline & \CC{offlinecolor}67.4 & \CC{offlinecolor}65.3 & \CC{offlinecolor}65.8 & \CC{offlinecolor}63.0 & \CC{offlinecolor}53.6 & \CC{offlinecolor}50.9 & \CC{offlinecolor}56.0 & \CC{offlinecolor}53.1 & \CC{offlinecolor}28.0 & \CC{offlinecolor}25.7 & \CC{offlinecolor}62.8 & \CC{offlinecolor}59.6 & \CC{offlinecolor}60.0 & \CC{offlinecolor}57.4 \\
\midrule

\multirow{2}[1]{*}{Supervised}   & \CC{ftcolor}Finetuning & \CC{ftcolor}63.0 & \CC{ftcolor}58.2 & \CC{ftcolor}56.9 & \CC{ftcolor}47.6 & \CC{ftcolor}49.1 & \CC{ftcolor}44.0 & \CC{ftcolor}55.7 & \CC{ftcolor}50.3 & \CC{ftcolor}27.7 & \CC{ftcolor}23.3 & \CC{ftcolor}68.6 & \CC{ftcolor}63.5 & \CC{ftcolor}55.9 & \CC{ftcolor}49.8 \\
                             \cmidrule{2-16}
                             & \CC{offlinecolor} Offline & \CC{offlinecolor}74.7  & \CC{offlinecolor}73.2 & \CC{offlinecolor}68.5  & \CC{offlinecolor}67.8 & \CC{offlinecolor}62.0  & \CC{offlinecolor}59.3 & \CC{offlinecolor}65.7  & \CC{offlinecolor}63.7 & \CC{offlinecolor}33.7  & \CC{offlinecolor}34.5 & \CC{offlinecolor}72.3  & \CC{offlinecolor}69.3 & \CC{offlinecolor}66.4  & \CC{offlinecolor}65.0 \\
\bottomrule
\end{tabular}
\end{table*}

The most similar to \name{} are data-focused regularization methods. Among them, a large majority leverage knowledge distillation using the outputs of a classifier learned with supervision \eg~\cite{Li17learning, castro2018end, fini2020online}, while a few works employ feature distillation~\cite{hou2019learning, douillard2020podnet} which is viable even without supervision. \cite{iscen2020memory} is also related to \name{}, but it focuses on memory efficiency which is less interesting in our setting. Also, \cite{iscen2020memory} explicitly uses the classifier after feature adaptation, hence it is unclear how to adapt it for CSSL, especially since in SSL positives are generated using image augmentations, which are not applicable to a memory bank of features. On the contrary, augmentations can be used in replay methods, among which we select the most common (ER~\cite{Robins95}) and one of the most recent (DER~\cite{buzzega2020dark}). Regarding prior-focused regularization methods, we choose EWC~\cite{kirkpatrick2017overcoming} over others (SI~\cite{Zenke17}, MAS~\cite{Aljundi17}, \etc) as it is considered the most influential and it works best with task boundaries. We also consider two CSSL baselines: LUMP~\cite{madaan2021rethinking} and Lin \etal~\cite{lin2021continual}. Finally, we do not consider methods based on VAEs~\cite{rao2019continual, achille2018life}, since they have been shown to yield poor performance in the large and medium scale. For instance, as found by~\cite{falcon2021aavae}, a VAE trained offline on CIFAR10 reaches an accuracy of 57.2\%, which is lower than any method (except VICReg) trained continually on CIFAR100 with \name{}.

\paragraph{Implementation.} For EWC, we use the SSL loss instead of the supervised loss to estimate importance weights. For POD and Less-Forget, we only re-implement the feature distillation without considering the parts of their methods that explicitly use the classifier. For DER, we replace the logits  of the classifier with the projected features in the buffer. We re-implement all these baselines by adapting them from the official implementation (POD), or from the Mammoth framework provided with~\cite{buzzega2020dark} (DER, ER, EWC), or from the paper (Less-Forget). We also compare with two concurrent works that propose approaches for CSSL (LUMP~\cite{madaan2021rethinking}, Lin \etal~\cite{lin2021continual}). LUMP uses k-NN evaluation, therefore we adapt the code provided by the authors to run in our code base. For Lin \etal, we compare directly with their published results, since they use the same evaluation protocol. We perform hyperparameter tuning for all baselines, searching over 5 values for the distillation loss weights of POD and Less-Forget, 3 values for the weight of the regularization in EWC and 3 replay batch sizes for replay methods. The size of the replay buffer is 500 samples for all replay based methods.

\section{Additional results}
\paragraph{Continual supervised contrastive with \name{}.} After the popularization of contrastive learning~\cite{chen2020simple, he2020momentum} for unsupervised learning of representations, \cite{khosla2020supervised} proposed a supervised version of the contrastive loss. Here, we show that \name{} is easily extendable to support supervised contrastive learning. The implementation is basically the same as for our vanilla contrastive-based distillation loss. In Tab.~\ref{tab:supervised-contrastive}, we show the improvement that \name{} brings with respect to fine-tuning, which is sizeable in the class-incremental setting. We also report the same comparison on DomainNet in Tab.~\ref{tab:domain-incremental-task-agnostic}, showing interesting results in both task-aware and task-incremental evaluation.

\begin{table}[t]
\caption{Linear evaluation top-1 accuracy on ImageNet100 (5 tasks, class- and data-incremental).}
\label{tab:supervised-contrastive}
\vspace{-2px}
\scriptsize
\centering
\captionsetup{type=table}
\begin{tabular}{lccc}
\toprule
\multirow{2}[1]{*}{\textbf{Method}} & \multirow{2}[1]{*}{\textbf{Strategy}} & \multicolumn{2}{c}{\textbf{ImageNet100}} \\ 
\cmidrule(lr){3-4}
&& Class-inc. & Data-inc. \\
\midrule
\multirow{2}{*}{{\parbox{1.2cm}{Supervised Contrastive}}}      & \CC{ftcolor}Fine-tuning & 61.6 & \CC{ftcolor}74.3 \\
                             & \CC{contrcolor}\name{} 
                             & \CC{contrcolor}\textbf{69.6}& \CC{contrcolor}\textbf{76.9}  \\ 
\bottomrule
\end{tabular}
\captionsetup{width=.99\linewidth}
\end{table}

\paragraph{Task-agnostic evaluation and domain-wise accuracy on DomainNet.} In the main manuscript, we showed that \name{} significantly improved performance in the domain-incremental setting using task-aware evaluation. Here, ``task-aware'' refers to the fact that linear evaluation is performed on each domain separately, \ie a different linear classifier is learned for each domain. However, it might also be interesting to check the performance of the model when the domain is unknown at test time. For this reason, we report the performance of our model when evaluated in a task-agnostic fashion. In addition, we also show the accuracy on each task (\ie domain). All this information is presented in Tab.~\ref{tab:domain-incremental-task-agnostic}. \name{} \textbf{always} outperforms fine-tuning with both evaluation protocols. The accuracy of \name{} on ``Clipart'' is also higher than offline. This is probably due to a combination of factors: (i) Clipart is the last task, therefore it probably benefits in forward transfer and (ii) a similar effect to the one found in~\cite{tian2021divide}, where dividing data in subgroups tends to enable the learning of better representations. Also, we notice that task-agnostic accuracy is lower than the task-aware counterpart. This is expected and means that the class conditional distributions are not perfectly aligned in different domains. As in the main paper, the colors are related to the type of SSL loss.

\begin{table}[t]
\caption{k-NN evaluation on ImageNet100 (5 tasks, class-incremental) performed on backbone and projected features.}
\label{tab:knn}
\vspace{-2px}
\scriptsize
\centering
\captionsetup{type=table}
\begin{tabular}{lccc}
\toprule
\multirow{2}[1]{*}{\textbf{Method}} & \multirow{2}[1]{*}{\textbf{Strategy}} & \multicolumn{2}{c}{\textbf{k-NN accuracy ($\uparrow$)}} \\
\cmidrule(lr){3-4}
&& \textbf{Backbone ($f_b$)} & \textbf{Projector ($f_p$)} \\
\midrule
\multirow{2}{*}{{\parbox{0.8cm}{Barlow Twins}}}      & \CC{ftcolor}Fine-tuning & 59.1 & \CC{ftcolor}34.4 \\
                             & \CC{decorrcolor}\name{} 
                             & \CC{decorrcolor}\textbf{63.4}& \CC{decorrcolor}\textbf{53.2}  \\ 
\midrule
\multirow{2}{*}{SwAV}      & \CC{ftcolor}Fine-tuning & \textbf{60.0} & \CC{ftcolor}53.9 \\
                             & \CC{knowcolor}\name{} 
                             & \CC{knowcolor}59.7 & \CC{knowcolor}\textbf{61.3}  \\ 
\midrule
\multirow{2}{*}{BYOL}     & \CC{ftcolor}Fine-tuning & 57.1 & \CC{ftcolor}33.0 \\
                             & \CC{predcolor}\name{} 
                             & \CC{predcolor}\textbf{61.2}& \CC{predcolor}\textbf{60.8}  \\ 
\midrule
\multirow{2}{*}{VICReg}       & \CC{ftcolor}Fine-tuning & 56.7 & \CC{ftcolor}35.3 \\
                             & \CC{predcolor}\name{} 
                             & \CC{predcolor}\textbf{59.5}& \CC{predcolor}\textbf{43.4}  \\ 
\midrule
\multirow{2}{*}{MoCoV2+}      & \CC{ftcolor}Fine-tuning & 54.5 & \CC{ftcolor}39.0 \\
                             & \CC{contrcolor}\name{} 
                             & \CC{contrcolor}\textbf{61.5}& \CC{contrcolor}\textbf{53.1}  \\ 
\midrule
\multirow{2}{*}{SimCLR}      & \CC{ftcolor}Fine-tuning & 54.8 & \CC{ftcolor}40.1 \\
                             & \CC{contrcolor}\name{} 
                             & \CC{contrcolor}\textbf{61.7}& \CC{contrcolor}\textbf{53.2}  \\ 
\bottomrule
\end{tabular}
\captionsetup{width=.99\linewidth}
\vspace{-8px}
\end{table}

\paragraph{Additional results with k-NN evaluation.}
For completeness, in this supplementary material, we also show that \name{} yields superior performance when evaluated with a k-NN classifier instead of linear evaluation. We use weighted k-NN with l2-normalization (cosine similarity) and temperature scaling as in~\cite{caron2021emerging}. Since since k-NN is much faster than linear evaluation we could also assess the quality of the projected representations, instead of just using the backbone. The results can be inspected in Tab.~\ref{tab:knn}. Three interesting phenomena arise: (i) \name{} always improves with respect to fine-tuning, (ii) the features of the backbone $f_b$ are usually better than the features of the projector $f_p$ and (iii) \name{} causes information retention in the projector, which significantly increases the performance of the projected features. An exception is represented by SwAV~\cite{caron2020unsupervised}, that seems to behave differently to other methods. First, the accuracy of the projected features in SwAV is much higher than other methods. 
This might be due to the fact that it uses prototypes, which bring the representations 1 layer away from the loss, making them less specialized in the SSL task. Second, it seems that \name{} only improves the projected features when coupled with SwAV. However, this is probably an artifact of the evaluation procedure, as the l2-normalization probably causes loss of information. Indeed, although the overall performance is lower, SwAV + \name{} outperforms SwAV + fine-tuning (58.7\% vs 56.9\%) if the euclidean distance is used in place of the cosine similarity for the backbone features. We leave a deeper investigation of this phenomenon for future work.

\begin{table}[t]
\caption{Linear evaluation top-1 accuracy on CIFAR100 (10 tasks, class-incremental).}
\label{tab:10-tasks}
\scriptsize
\centering
\captionsetup{type=table}
\begin{tabular}{lccc}
\toprule
\textbf{Method} & \textbf{Strategy} & \textbf{A ($\uparrow$)} \\ 
\midrule
\multirow{2}{*}{SimCLR}      & Fine-tuning & 39.3 \\
                             & \CC{contrcolor}\name{} & \CC{contrcolor}\textbf{52.7} \\ 
\midrule
\multirow{2}{*}{Barlow Twins}  & \CC{ftcolor}Fine-tuning & 49.9\\                      
  & \CC{decorrcolor}\name{} & \CC{decorrcolor}\textbf{53.7} \\ 
\bottomrule
\end{tabular}
\captionsetup{width=.99\linewidth}
\end{table}

\begin{table}[t]
\caption{Linear evaluation top-1 accuracy on ImageNet100 (5 tasks, class- and data-incremental) with ResNet50~\cite{he2016deep}.}
\label{tab:r50}
\scriptsize
\centering
\captionsetup{type=table}
\begin{tabular}{lccc}
\toprule
\multirow{2}[1]{*}{\textbf{Method}} & \multirow{2}[1]{*}{\textbf{Strategy}} & \multicolumn{2}{c}{\textbf{A ($\uparrow$)}} \\ 
\cmidrule(lr){3-4}
&& \textbf{Class-inc.} & \textbf{Data-inc.} \\
\midrule
\multirow{2}{*}{SimCLR}      & Fine-tuning & 70.7 & 75.6 \\
                             & \CC{contrcolor}\name{} & \CC{contrcolor}\textbf{74.0} & \CC{contrcolor}\textbf{77.2}  \\ 
\midrule
\multirow{2}{*}{Barlow Twins}  & \CC{ftcolor}Fine-tuning & 71.2 & 75.8 \\                      
  & \CC{decorrcolor}\name{} & \CC{decorrcolor}\textbf{74.8} & \CC{decorrcolor}\textbf{78.1} \\ 
\bottomrule
\end{tabular}
\captionsetup{width=.99\linewidth}
\end{table}

\paragraph{Different number of tasks.} The analysis of CSSL settings that we show in the main manuscript is limited to the 5 task scenario. However, it is interesting to run the same benchmarks with a longer task sequence. Nonetheless, one should also remember that SSL methods are data hungry, hence the less data is available per task, the higher the instability of the SSL models. In Tab.~\ref{tab:10-tasks}, we present additional results with 10 tasks on CIFAR100 (class-incremental). Barlow Twins seems to hold up surprisingly well, finishing up at roughly 50\% accuracy, while SimCLR suffers in the low data regime. Nonetheless, \name{} outperforms fine-tuning with Barlow Twins, and to a very large extent with SimCLR. 
\paragraph{Deeper architectures.} The experiments we propose in the main manuscript feature a ResNet18 network. This is a common choice in CL. However, in SSL, it is more common to use ResNet50. For this reason, in Tab.~\ref{tab:r50} we show that the same behavior observed with smaller networks is also obtained with deeper architectures. More specifically, \name{} outperforms fine-tuning in both class- and data-incremental settings by large margins. 

\paragraph{The role of the predictor.} In the main manuscript, we provided an intuitive explanation of the role of the predictor network that maps the current feature space to the frozen feature space. This intuition is corroborated by extensive experimentation and ablation studies. However, one more thing that is worth mentioning is that the success of the predictor network might also be related to the findings in SimSiam~\cite{chen2021exploring}, BYOL~\cite{grill2020bootstrap} and DirectPred~\cite{tian2021understanding}. 
Moreover, we perform additional ablations on the design of CaSSLe's predictor for SimCLR on CIFAR100 (5 tasks): adding BatchNorm after the hidden layer does not make any difference in terms of performance, and removing the non-linearity only causes a 0.3\% drop in accuracy.

\begin{table}[t]
\caption{Combinations of SSL methods and distillation losses on CIFAR100 (class-incremental, 2 tasks).}
\label{tab:comb-methods-distill}
\scriptsize
\centering
\captionsetup{type=table}
\begin{tabular}{lccc}
\toprule
\textbf{Distillation Loss}         & \textbf{SimCLR} & \textbf{Barlow Twins} & \textbf{BYOL}\\ 
\midrule
 InfoNCE & \CC{contrcolor}\textbf{61.8} & 64.5 & 64.8 \\
 Cross-correlation  & 60.1 & \CC{decorrcolor}\textbf{67.2} & 65.8  \\ 
 MSE & 61.3  & 64.6 & \CC{predcolor}\textbf{66.7} \\ 
\bottomrule
\end{tabular}

\end{table}

\paragraph{Combinations of SSL methods and distillation losses.} For computational reasons, it was not feasible to perform experiments combing all SSL methods with all possible distillation losses. However, in Tab.~\ref{tab:comb-methods-distill} we provide a subset of the possible combinations to validate our strategy that uses the same SSL loss for distillation.

{\small
\bibliographystyle{ieee_fullname}
\bibliography{bib}
}

\end{document}